\documentclass[a4paper, 10pt, conference]{ieeeconf}      
\usepackage{FG2024}

\FGfinalcopy 

\IEEEoverridecommandlockouts                              
\usepackage{graphicx, subfigure}
\usepackage{hyperref, url}
\usepackage{algorithm} 
\usepackage{algpseudocode}
\usepackage{color, colortbl}
\usepackage{comment}
\usepackage{balance}
\definecolor{Gray}{gray}{0.9}

\usepackage{cite}

\title{\LARGE \bf
\textit{SDFD}: Building a Versatile Synthetic Face Image Dataset with Diverse Attributes
}

\author{\parbox{16cm}{\centering
    {\large Georgia Baltsou, Ioannis Sarridis, Christos Koutlis and Symeon Papadopoulos}\\
    {\normalsize
    Information Technologies Institute @ CERTH, Greece}
}}

\usepackage{fancyhdr}

\begin{document}

\ifFGfinal
\thispagestyle{empty}
\pagestyle{empty}
\else
\author{Anonymous FG2024 submission\\ Paper ID \FGPaperID \\}
\pagestyle{plain}
\fi
\maketitle
\thispagestyle{fancy}

\begin{abstract}

AI systems rely on extensive training on large datasets to address various tasks. However, image-based systems, particularly those used for demographic attribute prediction, face significant challenges. Many current face image datasets primarily focus on demographic factors such as age, gender, and skin tone, overlooking other crucial facial attributes like hairstyle and accessories. This narrow focus limits the diversity of the data and consequently the robustness of AI systems trained on them. This work aims to address this limitation by proposing a methodology for generating synthetic face image datasets that capture a broader spectrum of facial diversity. Specifically, our approach integrates a systematic prompt formulation strategy, encompassing not only demographics and biometrics but also non-permanent traits like make-up, hairstyle, and accessories. These prompts guide a state-of-the-art text-to-image model in generating a comprehensive dataset of high-quality realistic images and can be used as an evaluation set in face analysis systems. Compared to existing datasets, our proposed dataset proves equally or more challenging in image classification tasks while being much smaller in size.

\end{abstract}

\section{INTRODUCTION}

AI systems are typically trained on large scale datasets. However, if such datasets are not balanced and diverse, there is a risk of ending up with unfair and inaccurate AI systems \cite{parraga2023fairness}. As a result, AI researchers and developers must be careful when selecting training and evaluation data. This is crucial for creating more reliable AI systems.

A particular application domain that calls for careful dataset composition is that of AI facial image analysis. 
More specifically, both the training and test images should be sufficiently broad and diverse to cover the numerous ways in which faces naturally differ. In the majority of existing face image datasets \cite{grimmer2021generation, karkkainen2021fairface}, diversity is considered through the inclusion of faces from different demographic groups defined on the basis of age, gender and skin tone. However, these dimensions are not sufficient to capture the whole spectrum of facial variety. Previous studies showed that lack of diversity can lead to AI system failures \cite{terhorst2021comprehensive}. Face verification systems, for example, may fail due to various types of occlusion \cite{taskiran2020face, oloyede2020review, zeng2021survey}. This observation motivated us to rethink of important dimensions in facial appearance and include dimensions related to different hair colors and styles, head and facial accessories or attributes like glasses, hats, tattoos, head-scarfs, and even hijabs, turbans, etc. We focus on creating an evaluation set since we believe it is a vital part of the AI system assessment process. 

In the present paper, we present a methodology for generating synthetic face image datasets with the goal of covering a wider range of the spectrum of facial diversity compared to existing ones. To this end, we move beyond demographics and biometrics to non-permanent characteristics like make-up, hair styles, accessories, etc. The proposed dataset creation methodology can be adjusted to the specific situation being examined. We present a particular instance of the proposed strategy, but AI researchers and practitioners may adapt it based on the particular context of use and task. Furthermore, we provide a new face image dataset, called \textit{SDFD}, which can be used as an evaluation set in AI systems. \textit{SDFD} comprises $1000$ different face images depicting people of diverse races, genders, ages, wearing a variety of accessories, different types of make-up and expressing various emotions.  Despite its relatively small size, the dataset captures a wide variety of different attributes and proves to be a challenging test set. 

The contributions of our study are twofold:
\begin{itemize}
    \item From a {\bf modeling perspective}, we suggest a methodology for generating realistic synthetic face image datasets that can be useful to AI systems evaluation. The current study contributes to the differentiation of existing datasets by taking into account additional face traits beyond demographics and biometrics, which result in covering a wider spectrum of real-world face variety.
    \item From a {\bf practical perspective}, we provide a first version of the synthetic face image dataset \textit{SDFD}, created via the proposed methodology. This dataset has been designed to be as inclusive as possible in order to assist evaluating computer vision systems with respect to minority groups and outliers.

\end{itemize}

\section{RELATED WORK}

The widespread use of AI technologies in different aspects of society has generated concerns about fairness and equity \cite{selbst2019fairness, mehrabi2021survey}. One area of concern is fair face analysis and recognition, which uses AI algorithms to identify, analyze, and recognize faces. Such processes should be designed to eliminate biases caused by factors such as poor representation in training data, algorithmic design decisions, or social prejudices.
Most existing studies have either focused on recognizing bias in AI systems \cite{buolamwini2018gender, raji2019actionable, amini2019uncovering}, or suggesting techniques to mitigate it \cite{wang2020mitigating, wang2020towards, dooley2024rethinking}. Creating face image datasets that represent faces of diverse groups can help reduce bias in face analysis systems as they
can easily represent various demographic groups \cite{alvi2018turning, merler2019diversity, karkkainen2021fairface, melzi2024synthetic}. The major studies in this topic will be presented next, followed by methods for image generation.

\subsection{Face Image Datasets}
Existing face image datasets may consist of real-world or synthetic faces. Nowadays, there is a plethora of real-world face image datasets such as CelebA \cite{liu2015faceattributes}, FFHQ \cite{karras2019style}, and LFW \cite{huang2008labeled}. However, real-world datasets suffer from some major problems. The first is about how most of these datasets were created. The images, in particular, were retrieved from the Internet, so no specific license was granted \cite{colbois2021use}. The second issue is that datasets are strongly biased toward specific attributes such as gender, race, or age. More precisely, since the majority of these images portray famous individuals, several of them depict e.g. white people with make-up and professional lighting \cite{kortylewski2019analyzing, melzi2024synthetic}. Consequently, such datasets do not fairly represent the real-world face distribution. The research community is aware of the former issues and significant efforts have been made to resolve them. DiF (Diversity in Faces) \cite{merler2019diversity} constitutes a face image dataset that is constructed to ensure diversity in age, gender, skin tone, a set of craniofacial ratios, location, and a measure of lighting. Another work in the same direction is FairFace \cite{karkkainen2021fairface}, a face image dataset that attempts to mitigate race, gender, and age bias.

Synthetic face image datasets are also explored in many works. In such datasets, the diversity issue can be mitigated by stating all attributes that, if depicted in it, would cover the real-world face variation prior to the image generation process. Furthermore, synthetic datasets have the advantage of reducing unintentional demographic bias caused by uneven representation of demographic groups in training data, since the quantity of images per category can be controlled \cite{serna2021insidebias,terhorst2021comprehensive,melzi2024synthetic}. An other advantage of AI-generated face images is that these days such images are so realistic that they cannot be separated from actual face photos, as a recent survey \cite{miller2023ai} reveals. Nevertheless, also works that focus on the generation of face images \cite{deng2020disentangled, colbois2021use, wood2021fake, qiu2021synface, grimmer2021generation, mekonnen2023balanced, bae2023digiface, kim2023dcface} run into some problems. First, several of these datasets include low-quality or non-realistic images. Another problem, which is similar to that of real-world datasets, is that image generation tools have been mostly trained on images retrieved from the Internet. This might lead to bias in image generation. Besides, most of the synthetic datasets still focus on variety solely on the basis of demographics and biometric features. The objective of the present work is to fill this gap by suggesting a method to generate a fair and realistic face image dataset. Table~\ref{tab:datasets} summarizes existing datasets along with some important characteristics.

\begin{table*}
\begin{center}
\caption{Datasets and their main characteristics.}
\label{tab:datasets}
    \begin{tabular}{p{3cm} p{3.4cm} p{1.7cm} p{1.7cm} p{1.8cm} p{3.8cm} }
      \hline
      \bf  Dataset & \bf source & \bf  \#Images & \bf  Unique Ids & \bf \#Labels/Image & \bf Source of Labels \\
      \hline
    CelebA \cite{liu2015faceattributes}             & Real (Internet)  & $\sim$200k & $\sim$10k  & 40 & Professional Labeling Company\\ 
    FFHQ \cite{karras2019style}             & Real (Flickr) & 70k & N/A & - & - \\ 
    LFW \cite{huang2008labeled}             & Real (Internet)  & $\sim$14k & $\sim$6k & 1 (identity) & Human annotators \\ 
    DiF \cite{merler2019diversity}             & Real (Flickr)  & $\sim$970k & N/A & 2 & Human annotators \\ 
    FairFace \cite{karkkainen2021fairface}             & Real (Flickr, Twitter, Web)  & $\sim$101k & N/A & 3 & Human annotators \\ 
     VGGFace2 \cite{cao2018vggface2}             & Real (Web)  & 3.31m & $\sim$9k & 1 (identity) & Human annotators \\
    BUPT  \cite{wang2021meta}             & Real (Web)  & 2m \& 38k & 1,3m \& 28k & - & - \\
     RFW \cite{wang2019racial}             & Real (\cite{FFT})  & $\sim$40k & $\sim$12k & 1 (identity) &  Human annotators \\ 
    Syn-Multi-PIE \cite{wood2021fake}             & Synthetic (StyleGAN2)  & 100k & 100k & 70 &  Automatic  Landmark Detection \\ 
    Syn-LFW \cite{qiu2021synface} (multiple)            & Synthetic (DiscoFaceGAN)  & multiple & multiple & 5 & MTCNN \cite{zhang2016joint} \\ 
    SymFace \cite{grimmer2021generation}           & Synthetic (StyleGAN)  & $\sim$77k & $\sim$26k & - & - \\ 
    SFace\cite{boutros2022sface} & Synthetic (StyleGAN2-ADA) & $\sim$634k & $\sim$11k & 1 (identity) & Human annotators \\
     \cite{mekonnen2023balanced}             & Synthetic (StyleGAN)   & $\sim$10k & N/A & 1 (race) & Downstream Models + Manually \\
     IDiff-Face\cite{boutros2023idiff} (multiple) & Synthetic (Diffusion Model) & 80k - 500k & 5k - 10k & - & - \\
    DigiFace-1M 1 \& 2 \cite{bae2023digiface}            & Synthetic (Cycles Rendering)  & 720k \& 500k & 10k \& 100k & - & - \\   
    DCFace 1 \& 2 \cite{kim2023dcface}             & Synthetic (DDPM)  & 500k \& 1m & 10k \& 20 & - & - 
       \\
       Arc2Face\cite{papantoniou2024arc2face} & Synthetic (SD v1.5) & 21m & 1m & - & -\\
      \hline
    \end{tabular}
    \end{center}
    \vspace{-5mm}
\end{table*}

\subsection{Image Generation Methods}
Text-to-image generation is an area that is rapidly progressing, with several studies appearing that lead to improved outcomes. As a result, many approaches have been proposed in the last years for generating images using text prompts. \cite{khan2020realistic}. These fall into four main categories \cite{zhan2023multimodal}: a) Generative Adversarial Networks (GAN) (e.g., StyleCLIP \cite{patashnik2021styleclip}, GigaGAN \cite{kang2023scaling}), b) autoregressive methods (e.g., Parti \cite{yu2022scaling}, Muse \cite{chang2023muse}), c) diffusion models (e.g., Imagen \cite{saharia2022photorealistic}, DALL-E2 \cite{ramesh2022hierarchical}, Stable Diffusion \cite{Rombach_2022_CVPR}, InstructPix2Pix \cite{brooks2023instructpix2pix}), and d) Neural Radiance Fields (NeRFs) (e.g., Magic3D \cite{lin2023magic3d}, ProlificDreamer \cite{wang2023prolificdreamer}). Besides, there are approaches that combine methods of the former categories like the GANDiffFace \cite{melzi2023gandiffface}, which combines GAN and Diffusion models.

Despite their very good results, GANs may suffer from \textit{mode collapse}, which causes the generator to produce restricted and repetitive outcomes, making them difficult to train. As stated in \cite{tomar2023review}, there are several works that attempt to alleviate this issue. Furthermore, these approaches require a significant amount of training data to perform well.

Autoregressive techniques may produce high-fidelity images, but there is no guarantee that the whole data distribution is captured. In addition, such methods suffer from slow inference speed.

Diffusion models allow users to adjust the quality and variety of generated images by providing precise control over the generating process \cite{ho2020denoising}. Although such methods can be computationally demanding, they produce remarkably realistic images. Creating realistic images is simply one aspect of an effective text-to-image model. Another key aspect to consider, is whether the output image matches the semantics of the input text description. Diffusion models are experimentally proven to have higher text-image alignment compared to other types of models discussed here \cite{kang2023scaling, lee2023holistic}. Also, like GAN methods, a substantial amount of training data is required for these approaches to perform successfully.

Finally, NeRF methods can result in 3D images, often of poor quality. In addition, NeRF approaches typically suffer from high computational costs.

\section{METHODOLOGY FOR GENERATING SYNTHETIC FACE IMAGE DATASETS}
\label{sec:process}

\begin{figure}[thpb]
\centering
\includegraphics[scale=0.17]{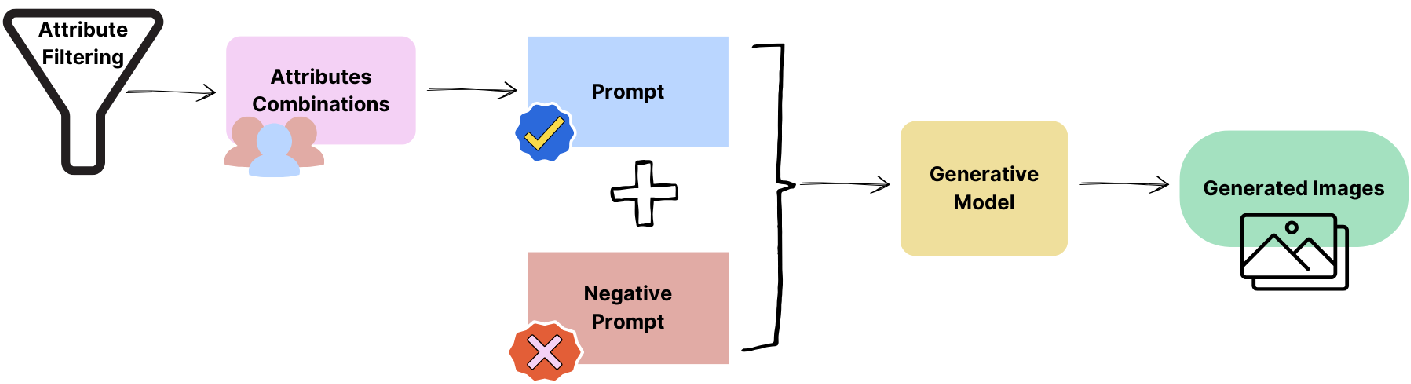}
\caption{Overview of proposed image generation process.}
\label{fig:process}
\end{figure}

Here, we present the general methodology that someone can follow in order to generate a diverse face image dataset. This involves a systematic process designed to produce diverse and realistic face images using the stable diffusion model, and is briefly illustrated in Figure~\ref{fig:process}. 

The following steps outline the proposed methodology:

\begin{enumerate}
  \item \textbf{Attribute collection and filtering.} The first step is to compile a list of terms that will represent the attributes that one would like to be depicted in the target face images. These terms should be carefully picked and then filtered to eliminate specific words or phrases.
  \item \textbf{Combinations of attributes.} After completing the basic attribute list, combinations of them should be created to generate the text prompt. 
  \item \textbf{Prompt formulation.} Prompts are used as input text to the generative model, describing the image that should be constructed. The more exact the description, the closer the resulting image will be to what was envisioned. Besides, negative prompts are also important since they specify what should not be included in the generated image. 
  \item \textbf{Generative model.} Prompts, together with negative prompts, act as input to the generative model, which produces the final images. This might be a either a diffusion model, or any other state-of-the-art model.
\end{enumerate}

The above steps might be adjusted depending on the context of use and task. For example, multiple attributes could be utilized, resulting in different prompts. The same is true for the generative model used.

It should be mentioned that the former image generation process may need to be performed several times before the user obtains the desired outcome. This, depends on the generative model used as well as the application.

In the next subsections, we discuss in further depth the steps that we followed to generate the proposed face image dataset and also the requirements we established.  

\subsection{Attribute Collection and Filtering}

This work's aim is to create a photo-realistic face image dataset that spans the whole range of real-world facial variability. To do this, a list of terms or phrases was compiled that characterize individuals of diverse gender, race, and age, as well as a variety of other characteristics. This list, together with a corresponding list of negative terms, provides the input text according to which images are created. 

Race options were selected based on the work of \cite{karkkainen2021fairface}. The race of \textit{Pacific Islanders} was also added, as it is considered a group with distinct characteristics that is not included in the former. The other attributes' options were empirically selected, using as basis previous research works \cite{samangouei2017facial, luccioni2024stable, microsoft, terhorst2021comprehensive}. More specifically, after many experiments with each word or phrase if the resulting image did not depict the input text sufficiently, the former was not used again. E.g., for the non-binary gender attribute $20$ different words or phrases were tested before concluding on the four options presented in Table~\ref{tab:prompts}.
All terms used to produce facial images are listed in Table~\ref{tab:prompts}.

\begin{table*}
\begin{center}
\caption{Terms used in prompts. Options are presented in alphabetical order.}
\label{tab:prompts}  
    \begin{tabular}{p{1.9cm} p{14.5cm} }
      \hline
      \bf  Attribute  & \bf  Options\\
      \hline                                                                                                                                             
gender           & androgynous person, baby, boy, gender fluid person, girl, man, person having an androgynous appeal/male and female characteristics, woman \\ \hline
\rowcolor{Gray} mood             & angry, depressed, excited, happy, sad, scared, smiled, stressed, tired                                                                                   \\ \hline
race             & Black, East Asian, Indian,  Latino, Middle Eastern, Pacific Islander, Southeast Asian, White                                                             \\ \hline
\rowcolor{Gray} hat              & cap, hat, headscarf, helmet, pamela hat                                                                                                                                                         \\ \hline
vision           & color contact lenses, glasses, sunglasses                                                                                                                                                         \\ \hline
\rowcolor{Gray} hair colour      & black, blonde, blue, brown, green, pink, purple, red, white                                                                                                                                                         \\ \hline
hair style       & bald, braid, curly, dreadlocks, hair clips, headband, long, short, straight                                                                                                                                                         \\ \hline
\rowcolor{Gray} facial hair      & beard, moustache                                                                                                                                                         \\ \hline
face paints      & heavy make-up, lipstick, tattoo                                                                                                                                                         \\ \hline
\rowcolor{Gray} other facial     & earrings, medical mask, wrinkles                                                                                                                                                         \\ \hline
religious item & al-amira, burqa, chado, ghoongha, hijab, khimar, kippah, niqab, nun hat, snood, tichel, turban, veil                                                                                                                                                         \\ \hline
\rowcolor{Gray} resolution         & 4K, 8K, Ultra HD                                                                                                                                                         \\ \hline
camera           & Canon Eos 5D, Fujifilm XT3, Nikon Z9, shot on iPhone                                                                                                                                                         \\ \hline
\rowcolor{Gray} pose             & front face, side profile                                                        
 \\
      \hline
\end{tabular}
 \end{center}  
 \vspace{-5mm}
\end{table*}

\subsection{Combinations of Attributes}

The next step is to create meaningful attribute combinations that may be utilized as the prompt basis. Concerning race, in order to incorporate as many various faces from all over the world as possible, combinations of the eight distinct race groups were also created at random, such as Black-Asian, Indian-White, etc.

Besides race, all the other attributes were combined in an almost random way to create as diverse face images as possible. The combination is not fully random, since there are some combinations that are not allowed due to their incompatibility. For example, \textit{hair style} attribute options except for \textit{bald}, could be combined with a \textit{hair color} attribute option. However, none option of \textit{hat} attribute can be combined with \textit{religious symbol} attribute options.

\subsection{Prompt Formulation}

Creating an appropriate prompt is a critical step in the acquisition of photo-realistic facial images \cite{papa2023use}. The attribute combinations formed in the previous step are utilized as a prompt, describing the key characteristics of the image to be generated. Aside from this, one should also define traits that should not be displayed in the final image. These are the negative prompt components.

In the present work, some restrictions concerning the (negative) prompts were also applied to generate as realistic images as possible. The following restrictions apply to each generated image in terms of (negative) prompting: (a) must not have the appearance of a computer generated or animated image, (b) must contain only a person's face in a realistic background, and (c) stereotypical traits with reference to any attribute (e.g. Indian person wearing traditional clothes and jewellery) should be eliminated. It was generally attempted to use as objective terms as possible to describe the faces, but there is a possibility that some stereotypical language may have been used. However, the goal was to leverage stereotypical elements to create a more diverse dataset.

Aside from these terms, some others were used in all of the experiments regardless of the attributes. The aim was to improve the realism of the generated images. These terms, hereafter called universal, are gathered in Table~\ref{tab:general_prompts}.

\begin{table}
\begin{center}
\small
\caption{Universal terms and negative terms used in prompts of all experiments.}
\label{tab:general_prompts}
\begin{tabular}{p{0.8cm} p{7.3cm} }
\hline
\textbf{Prompt} &
dry skin, realistic background, realistic dull skin noise, remarkable color, remarkable detailed pupils, skin fuzz, textured skin, tone mapping, ultra realistic, visible skin detail\\
\hline
\textbf{Negative Prompt} &
animated, bad anatomy, extra fingers, low resolution, unreal engine, worst quality\\ 
\hline

\end{tabular}
\end{center}
\vspace{-8mm}
\end{table}

The prompt formulation process is described in Algorithm~\ref{alg:1}. More precisely, each prompt is created by combining an attribute options list, the universal prompts and the negative prompts. The universal and negative prompts are consistent, as seen in Table~\ref{tab:general_prompts}. The attribute options list is generated in a random way, by selecting one, more or even none option for each attribute as described in the Table~\ref{tab:prompts}.

\begin{algorithm}[ht]
\caption{Prompt Formulation} 
	\begin{algorithmic}[1]
        \State $p = [ ]$
        \State $a_{optList} = [ ]$
		\For {$a \in attributes$ }
            \State $a_{opt} = random(a)$
            
            \State $a_{optList}.append(a_{opt})$
			
		\EndFor
            \State $p.append(a_{optList}) + p.append(p_{uni}) + p.append(p_{neg})$
	\end{algorithmic} 
\label{alg:1}
\end{algorithm}

\subsection{Diffusion Process}

\begin{figure}
\centering
\subfigure[][]{
\label{fig:oth-a}
\includegraphics[height=0.8in]{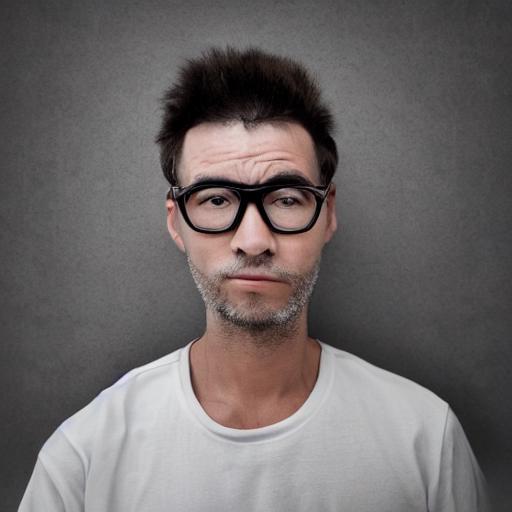}}
\subfigure[][]{
\label{fig:oth-b}
\includegraphics[height=0.8in]{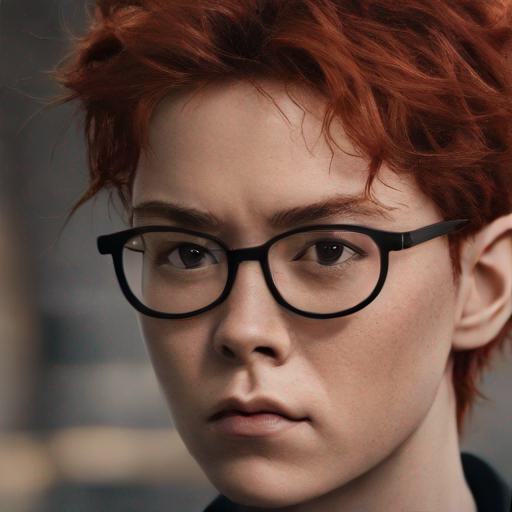}}
\caption[Examples of generated images with other stable diffusion versions]{Examples of generated images with other stable diffusion versions and their corresponding prompts without including universal prompt terms:
\subref{fig:oth-a} Stable Diffusion $1.5$: White, tired, man, wearing glasses, front face, Ultra HD, Nikon Z9,
\subref{fig:oth-b} Stable Diffusion xl Turbo: Black, angry, red hair, androgynous person, wearing glasses, side profile, Fujifilm XT3.}
\label{fig:otherModels}
\vspace{-5mm}
\end{figure}

We selected a Denoising Diffusion Probabilistic Model (DDPM) for text-to-image generation. Diffusion models have been shown to outperform others like GANs on high-quality image generation \cite{dhariwal2021diffusion, perera2023analyzing, chen2023face}. These models have strong semantic knowledge on many concepts and have been trained on a wide variety of images, including faces, through language supervision. In particular, the pretrained Stable Diffusion (SD) version $2.1$ \cite{Rombach_2022_CVPR} was used. SD versions $1.5$ and $xl$ were also tested, however the quality of generated face images were empirically found to be inferior to that of version $2.1$, as shown in Figures~\ref{fig:otherModels}. In general, the model works by breaking down the image generation process into various denoising steps. The process involves adding random noise to an image and gradually removing it in denoising steps to generate a final image. This diffusion process includes two phases: forward and backward. In the forward process, an image is transformed into noise through multiple steps, and in the backward process, the original image is restored by training a neural network to denoise a given noisy input. The model attempts to forecast the input of the forward process at each step.

\begin{figure}
\centering
\subfigure[][]{
\label{fig:inf-a}
\includegraphics[height=0.8in]{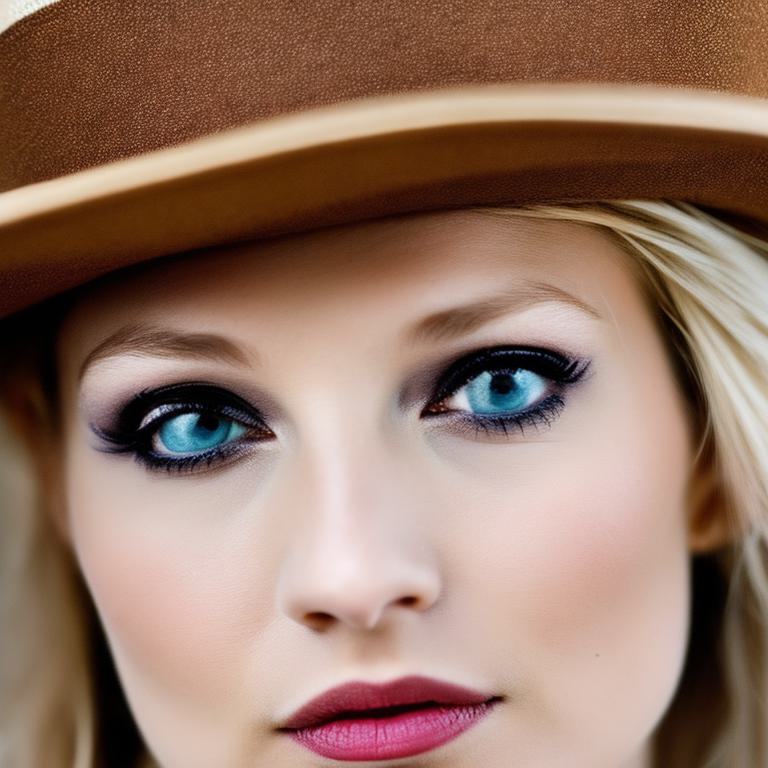}}
\subfigure[][]{
\label{fig:inf-b}
\includegraphics[height=0.8in]{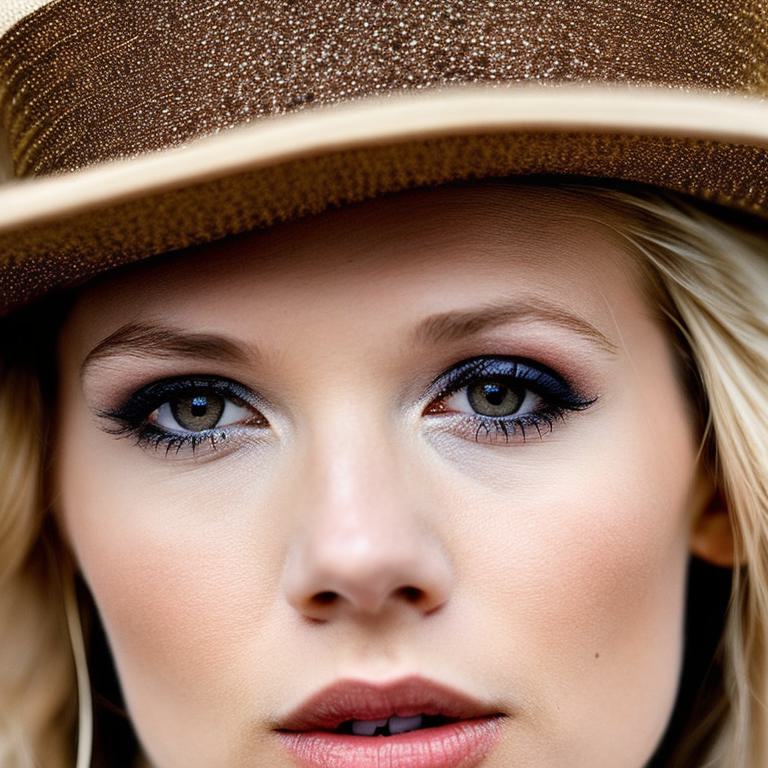}}
\subfigure[][]{
\label{fig:inf-c}
\includegraphics[height=0.8in]{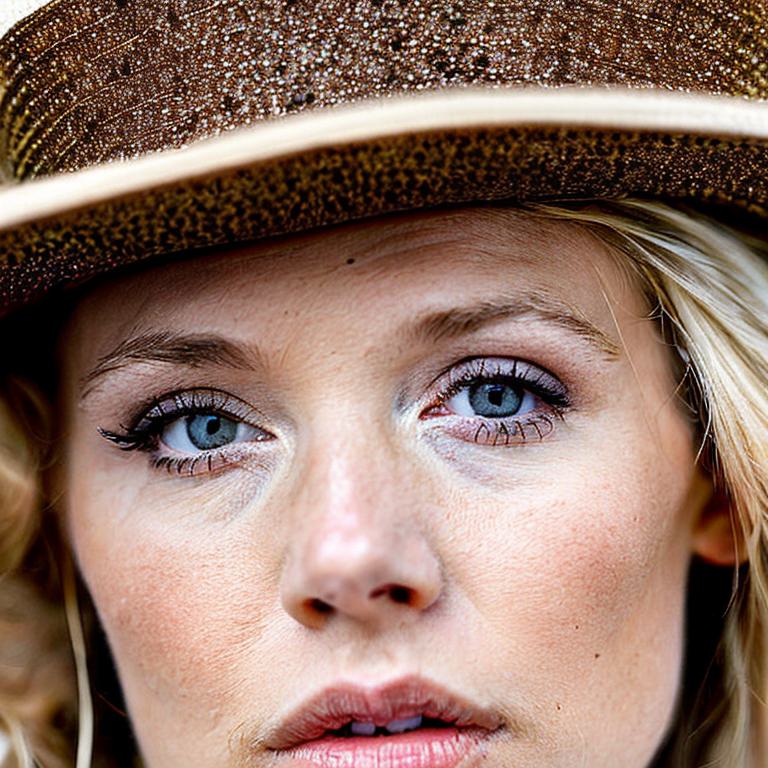}}
\subfigure[][]{
\label{fig:inf-d}
\includegraphics[height=0.8in]{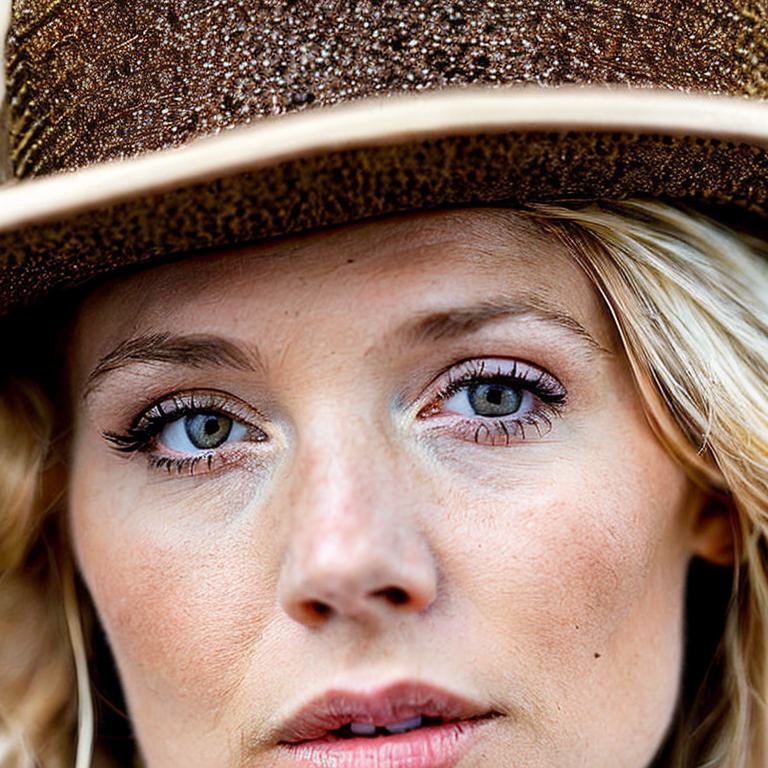}}
\subfigure[][]{
\label{fig:inf-e}
\includegraphics[height=0.8in]{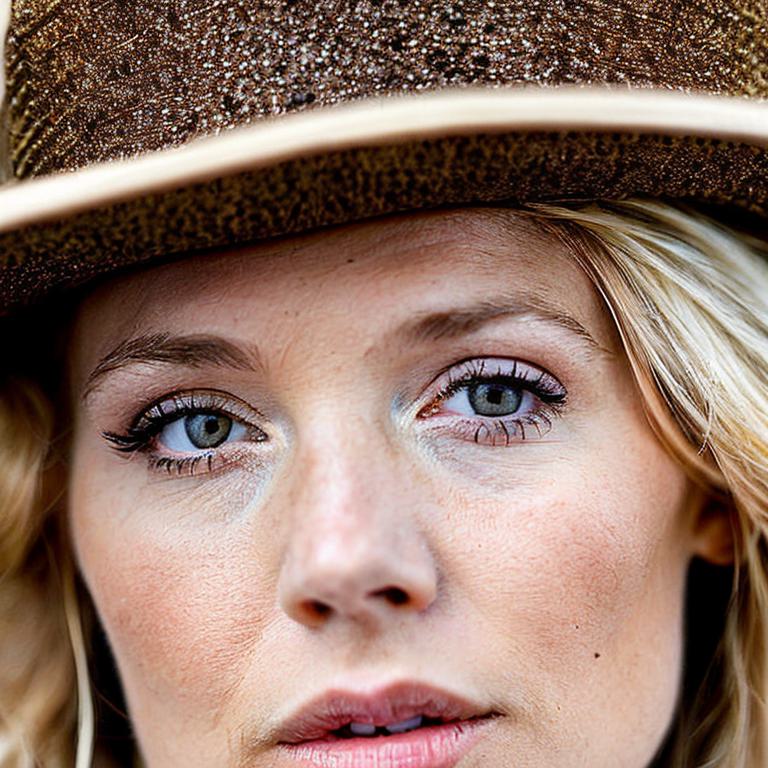}}
\subfigure[][]{
\label{fig:inf-f}
\includegraphics[height=0.8in]{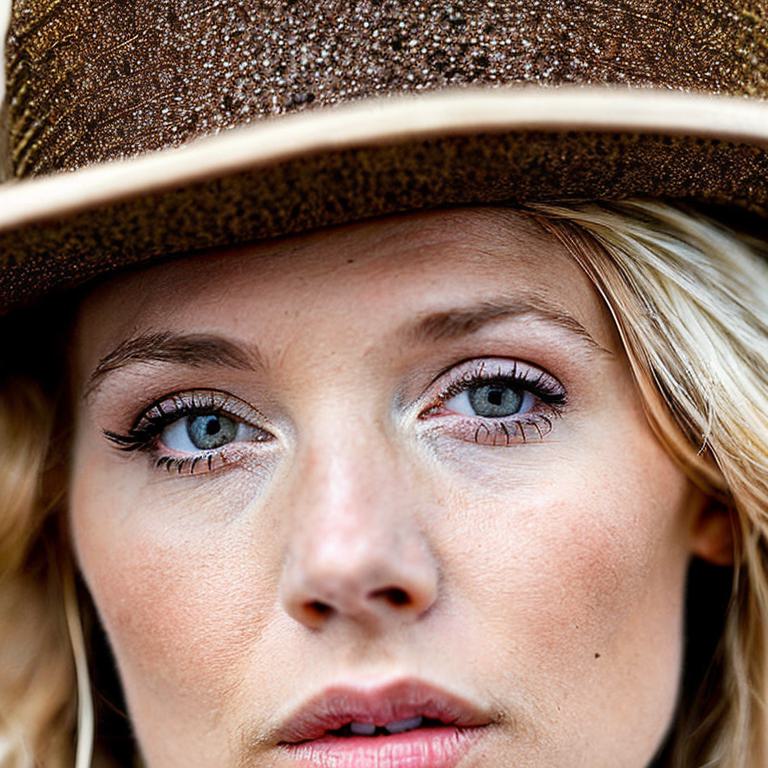}}
\caption[]{Example generated images with different values of inference steps for the prompt (apart from universal) "White, blonde woman, blue eyes, wearing hat":
\subref{fig:inf-a} 5 steps,
\subref{fig:inf-b} 10 steps,
\subref{fig:inf-c} 15 steps,
\subref{fig:inf-d} 30 steps,
\subref{fig:inf-e} 50 steps and
\subref{fig:inf-f} 70 steps.}
\label{fig:hyper1}
\vspace{-5mm}
\end{figure}

\begin{figure}
\centering
\subfigure[][]{
\label{fig:cfg-a}
\includegraphics[height=0.8in]{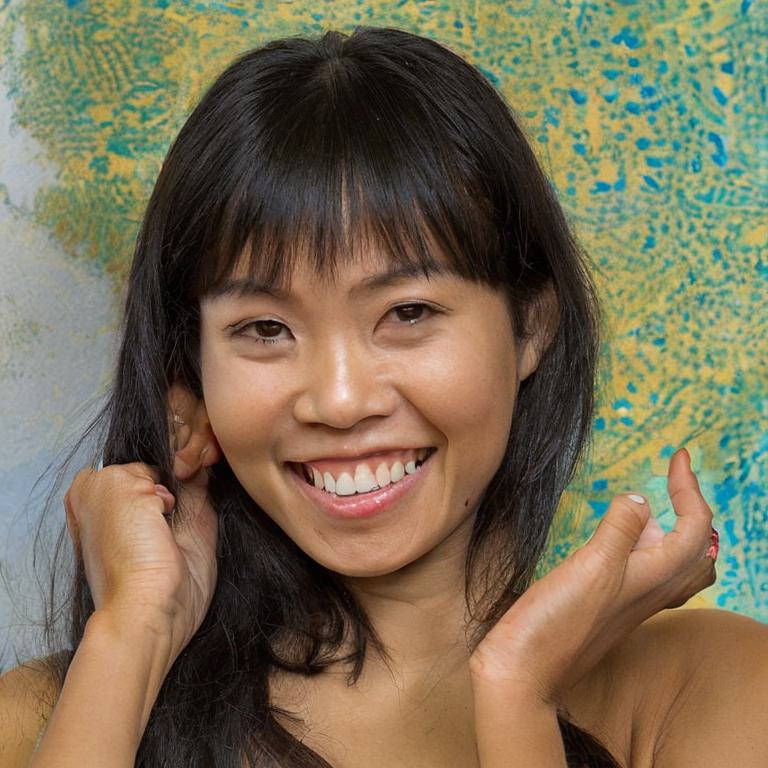}}
\subfigure[][]{
\label{fig:cfg-b}
\includegraphics[height=0.8in]{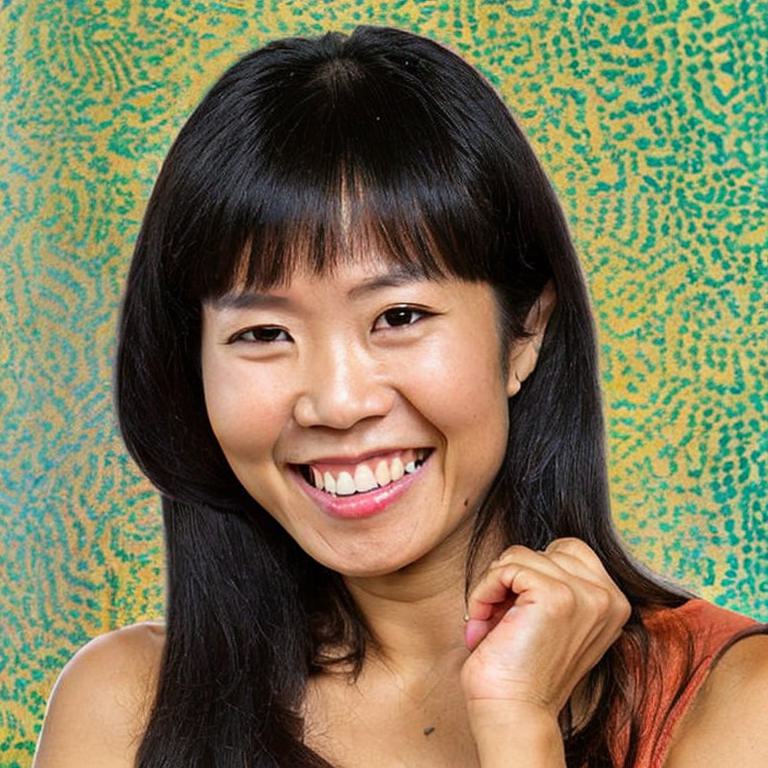}}
\subfigure[][]{
\label{fig:cfg-c}
\includegraphics[height=0.8in]{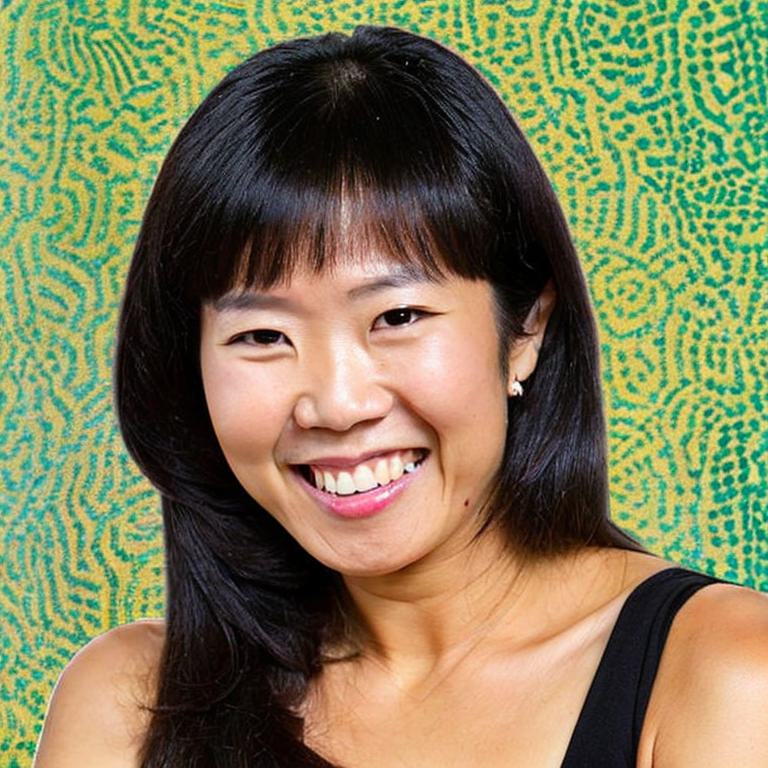}}
\subfigure[][]{
\label{fig:cfg-d}
\includegraphics[height=0.8in]{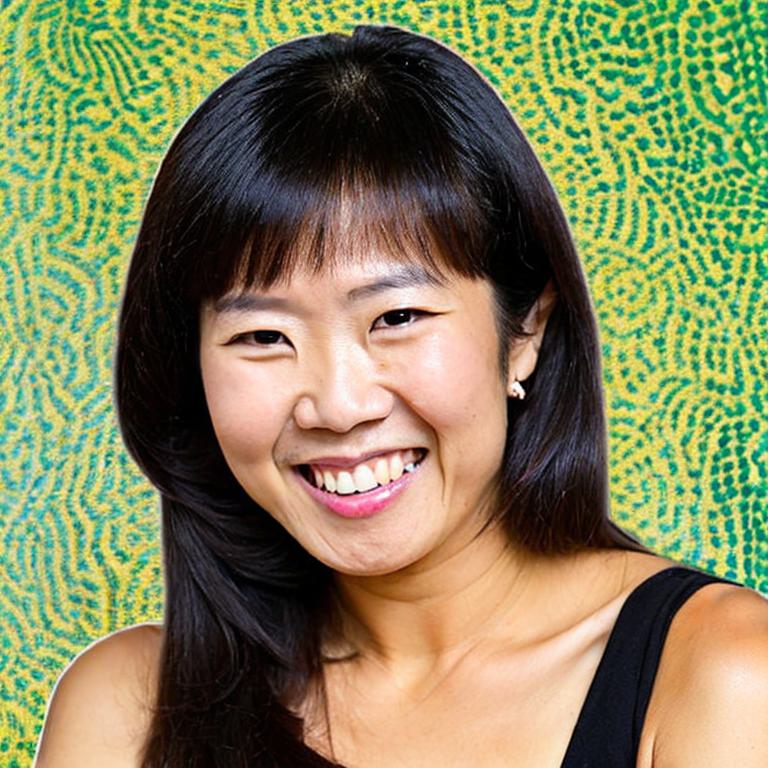}}
\subfigure[][]{
\label{fig:cfg-e}
\includegraphics[height=0.8in]{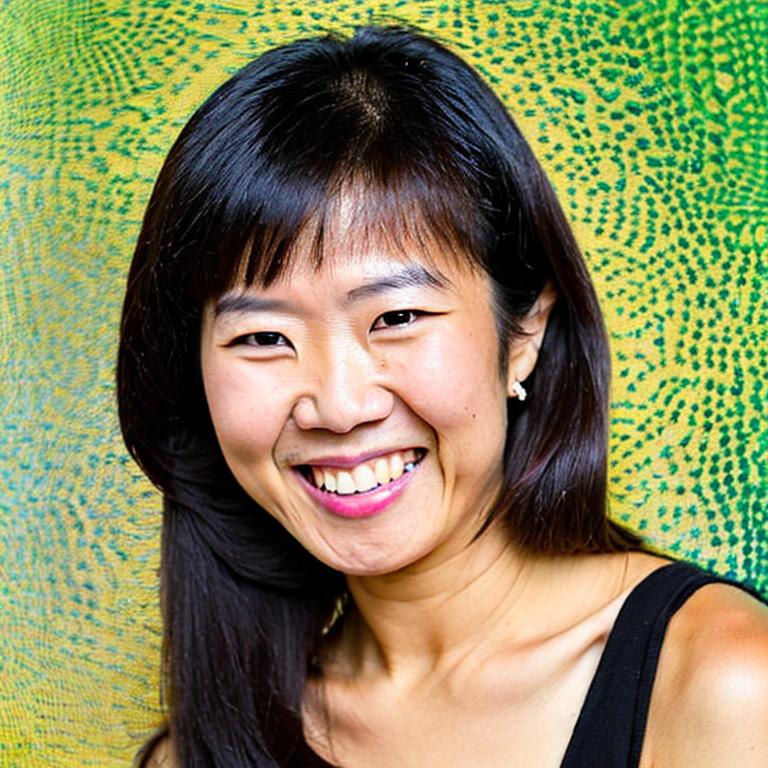}}
\subfigure[][]{
\label{fig:cfg-f}
\includegraphics[height=0.8in]{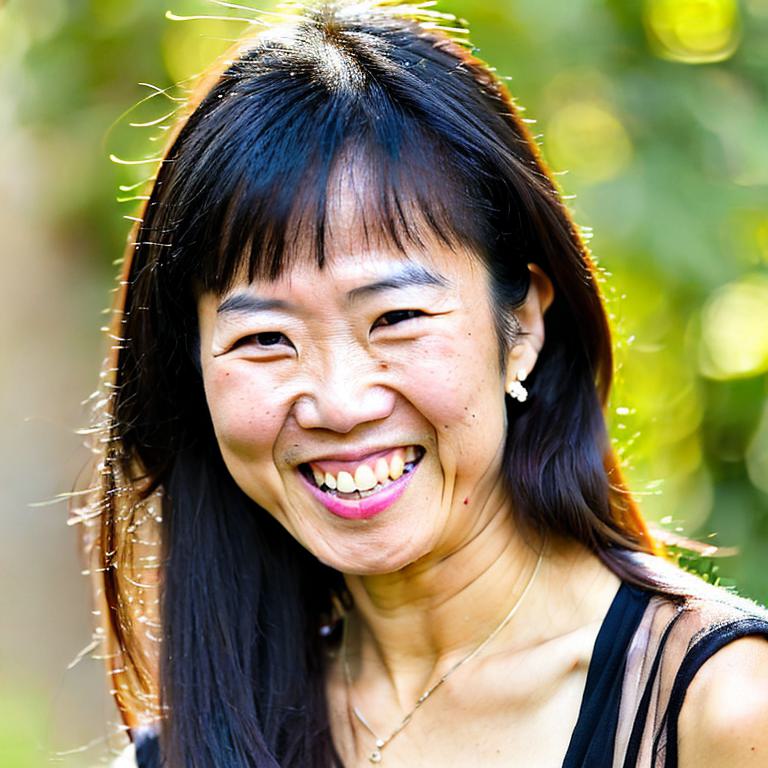}}
\caption[]{Example generated images with different values of CFG weight for the prompt (apart from universal) "Asian, woman, black hair, smiling":
\subref{fig:cfg-a} weight = 2.5,
\subref{fig:cfg-b} weight = 5,
\subref{fig:cfg-c} weight = 7.5,
\subref{fig:cfg-d} weight = 10,
\subref{fig:cfg-e} weight = 12.5 and
\subref{fig:cfg-f} weight = 15.}
\label{fig:hyper2}
\vspace{-5mm}
\end{figure}

The scheduler is an important parameter when utilizing a diffusion model since it defines the whole denoising process: the number of diffusion steps, how much noise is added on each step, etc. There is a plethora of different schedulers in the literature \cite{liu2021pseudo, lu2022dpm, song2020denoising, zhang2022fast} each trying to make the diffusion process more efficient while preserving good image quality. For the current work, the DPM-Solver$++$ \cite{lu2022dpm} was selected as it offers possibly the best speed-quality trade-off, with the number of inference (denoising) steps equal to $50$. Typically, the more steps, the better the result, but the longer the creation takes. Additionally, a Classifier Free Guidance (CFG) with a weight of $7.5$ was used. In general, the CFG weight affects the generator model's amount of flexibility when generating images, since the higher the weight, the greater the level of control by the provided prompt. For stable diffusion, values between $7$ and $8.5$ are typically appropriate selections \cite{valevski2023face0, rost2023stable}. Figures~\ref{fig:hyper1}, ~\ref{fig:hyper2} show how different values for inference steps and CFG weight respectively, affect the resulting image.

\begin{figure}
\subfigure[][]{
\label{fig:ex-a}
\includegraphics[height=1in]{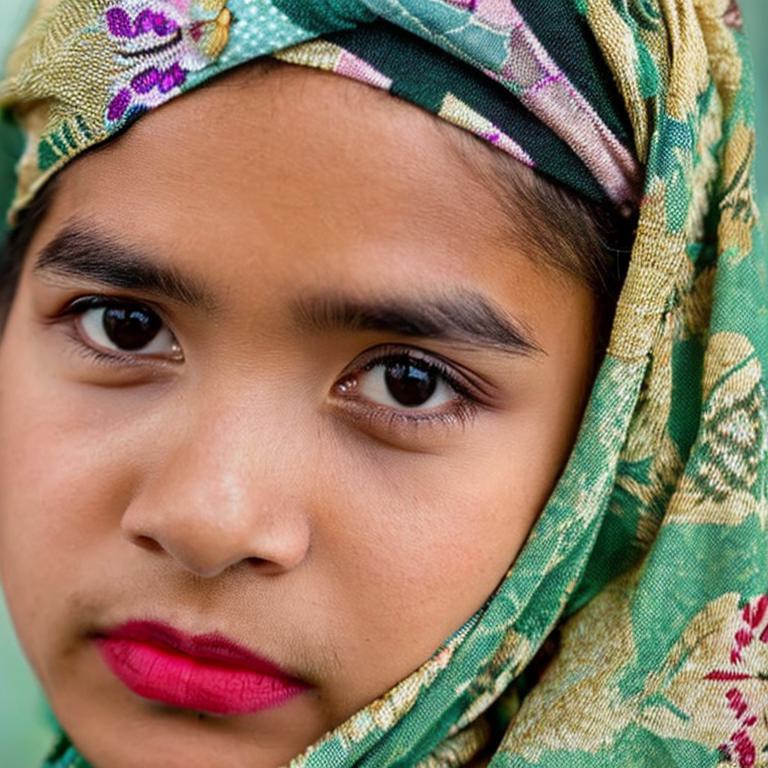}}
\subfigure[][]{
\label{fig:ex-b}
\includegraphics[height=1in]{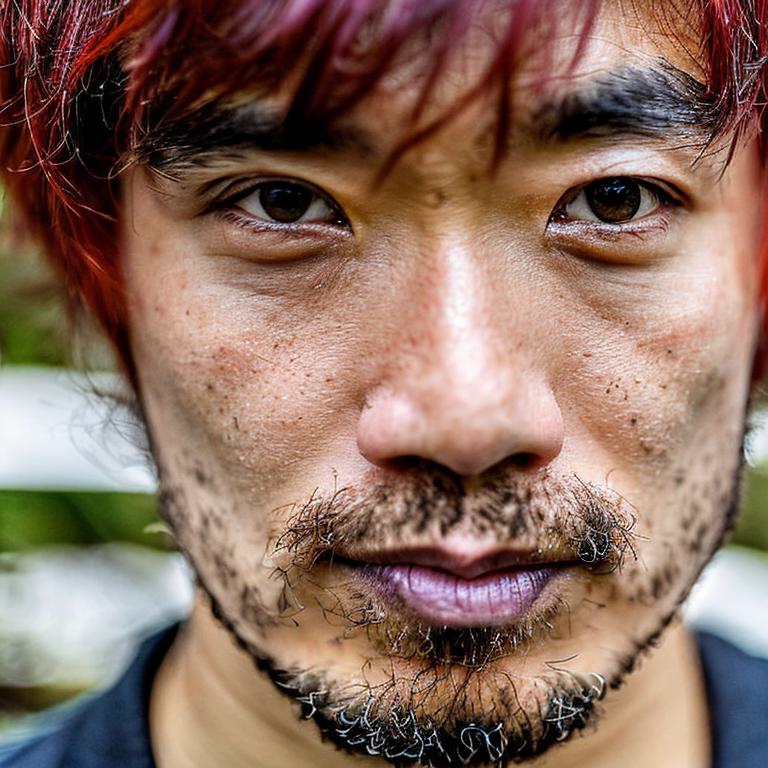}}
\subfigure[][]{
\label{fig:ex-c}
\includegraphics[height=1in]{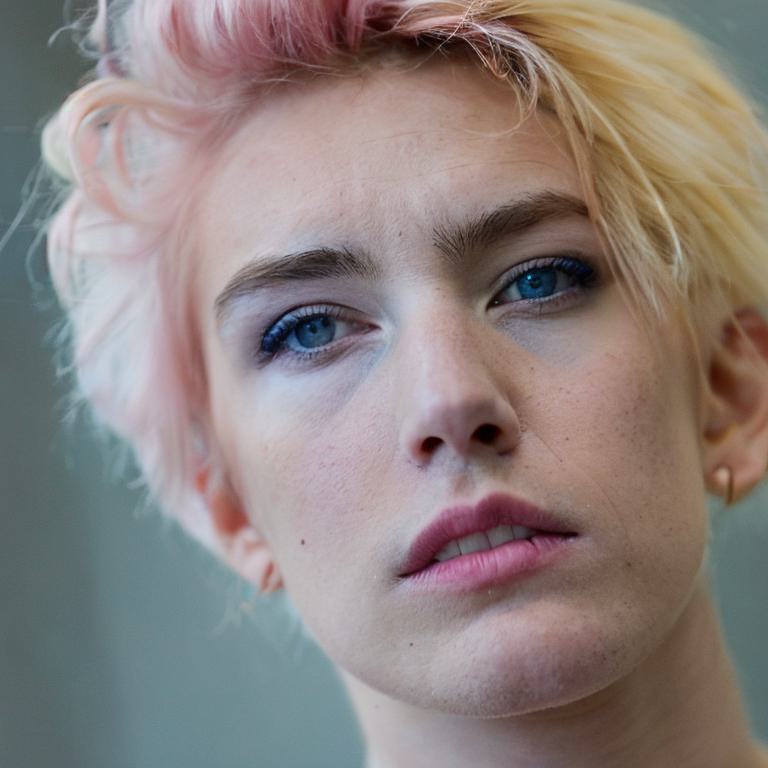}}
\caption[Example generated images]{Example generated images and their corresponding prompts without including universal prompt terms:
\subref{fig:ex-a} Pacific Islander, stressed, wearing headscarf, girl, black eyes, wearing lipstick,   8K,
\subref{fig:ex-b} East Asian, angry, red hair, man, wearing colour contact lenses, front face, Fujifilm XT3 and,
\subref{fig:ex-c} White, androgynous person, pink hair, blue eyes, Fujifilm XT3.}
\label{fig:examples}
\vspace{-4mm}
\end{figure}

\begin{figure}[ht]
\centering
\subfigure[][]{
\label{fig:in-a}
\includegraphics[height=1in]{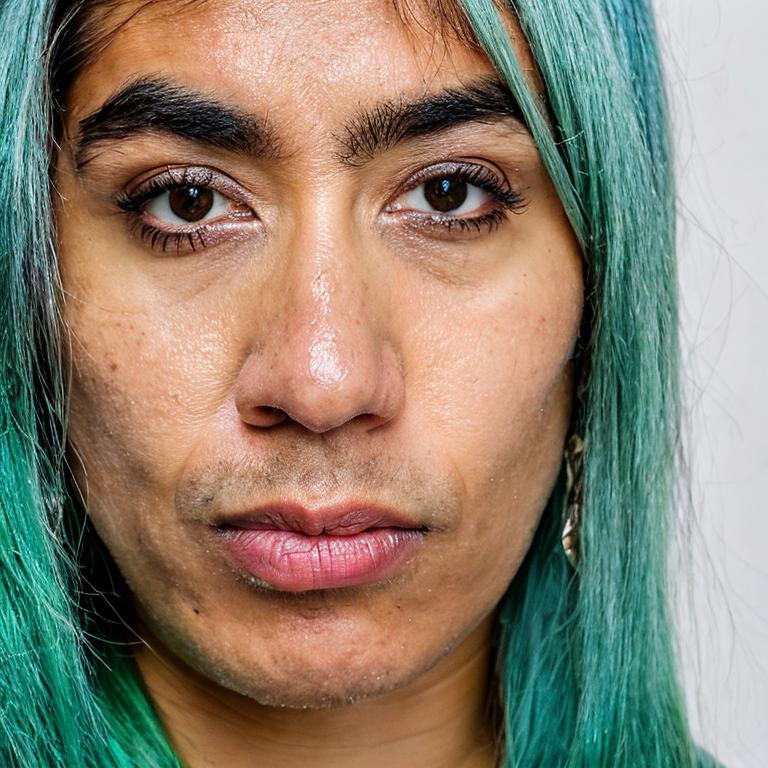}}
\subfigure[][]{
\label{fig:in-b}
\includegraphics[height=1in]{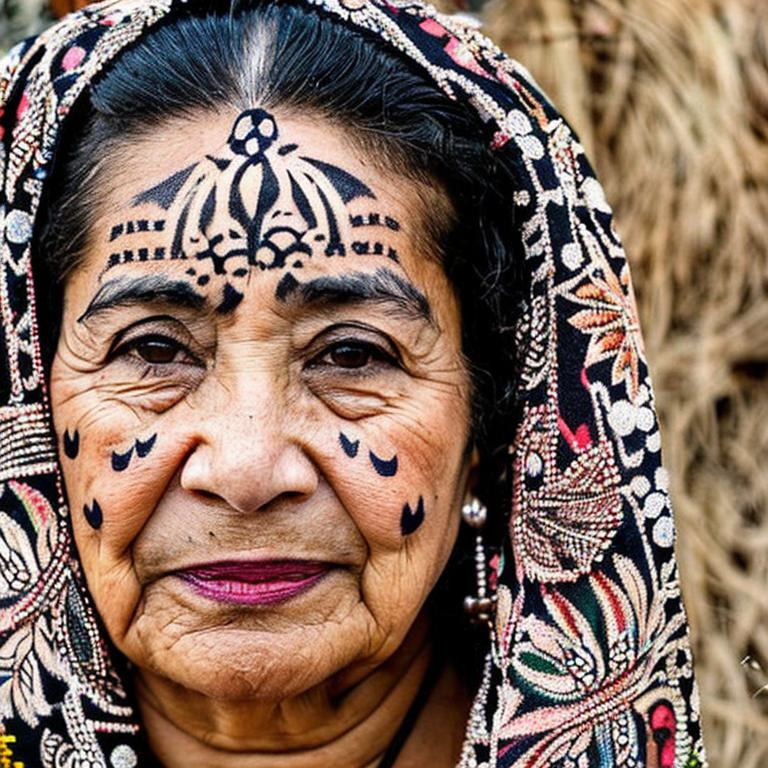}}
\subfigure[][]{
\label{fig:in-c}
\includegraphics[height=1in]{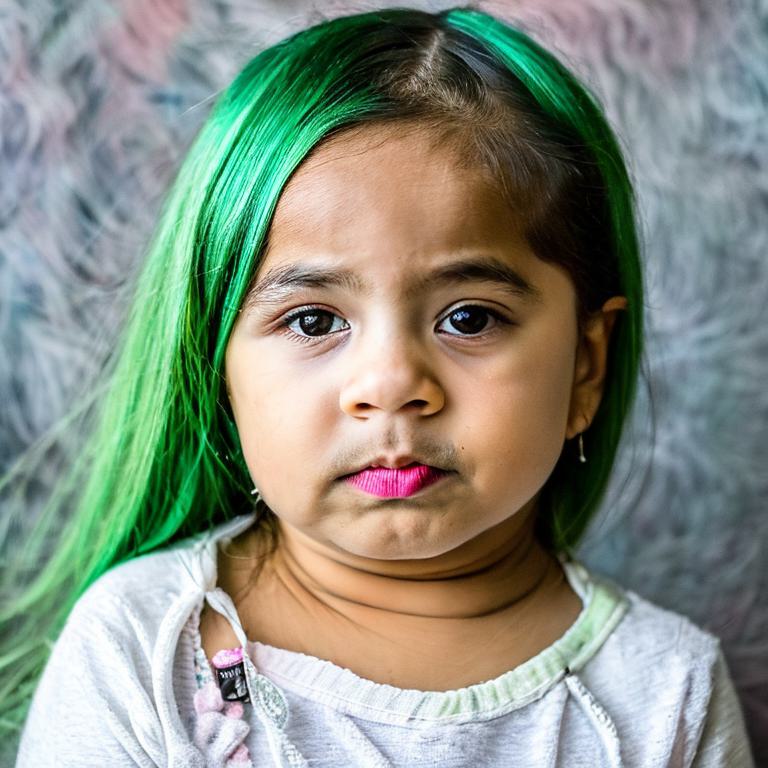}}
\caption[Examples of some interesting generated images]{Examples of some interesting generated images and their corresponding prompts except for universal ones:
\subref{fig:ex-a} Latino, stressed, green hair, person having male and female characteristics, front face, Fujifilm XT3,
\subref{fig:ex-b} Pacific Islander-Middle Eastern woman, having face tattoo, headband, having wrinkles,  8K, Nikon Z9 and,
\subref{fig:ex-c} Latino, baby girl, green hair, beard, black eyes, Fujifilm XT3.}
\label{fig:interesting}
\vspace{-5mm}
\end{figure}

Example generated face images are shown in Figure~\ref{fig:examples}.
The recommended random method of prompt formulation yields intriguing images that may be difficult to obtain in another way. As a result, the dataset's face diversity is boosted. Figure~\ref{fig:interesting} shows some examples of such images. 

\begin{figure}
\centering
\subfigure[][]{
\label{fig:age-a}
\includegraphics[height=1in]{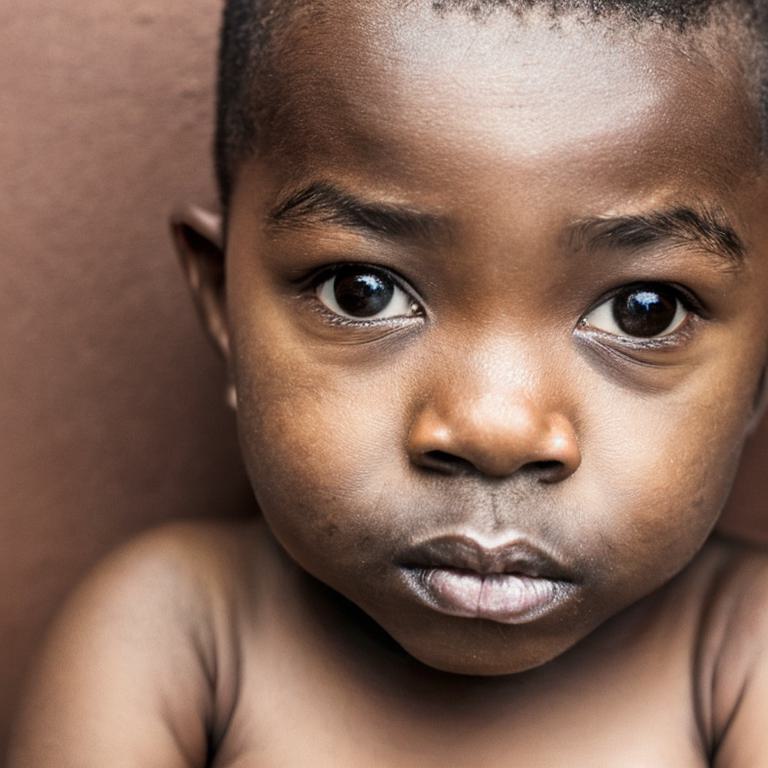}}
\subfigure[][]{
\label{fig:age-b}
\includegraphics[height=1in]{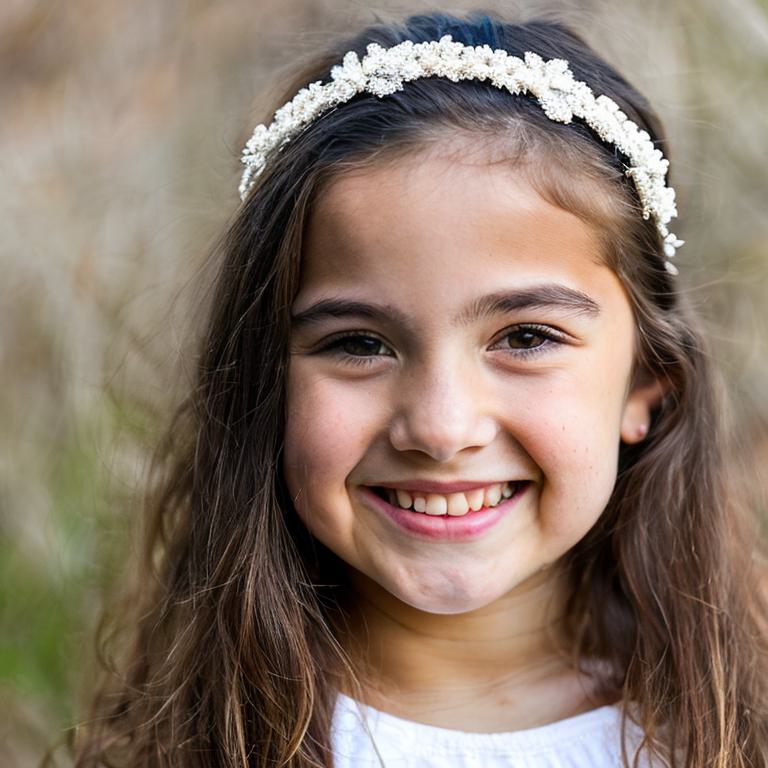}}
\subfigure[][]{
\label{fig:age-c}
\includegraphics[height=1in]{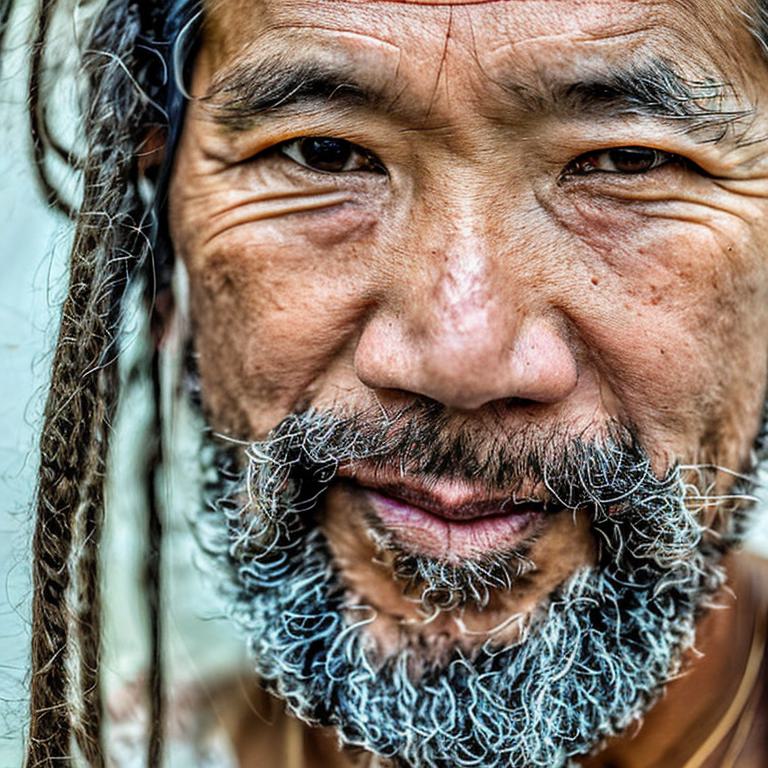}}
\caption[Example age generated images]{Example generated images and their corresponding prompts except for universal ones, depicting different age groups:
\subref{fig:age-a} Black, baby boy, short hair, 4K, Fujifilm XT3, 
\subref{fig:age-b} White, girl, headband, Ultra HD,  Nikon Z9,
\subref{fig:age-c} Southeast Asian-White, man, having piercing, dreadlocks, having wrinkles, Ultra HD, Canon Eos 5D.}
\label{fig:age}
\vspace{-5mm}
\end{figure}

It should be emphasized that the ``age'' attribute is implicitly applied through the attributes of gender, hair colour and other facial attributes. The gender attributes baby, boy and girl directly indicate the age range of a person. In addition, the terms ``white hair" and ``wrinkles" allude to an older person. Figure~\ref{fig:age} displays some generated face images that illustrate the age range of the dataset.

\begin{figure}
\centering
\subfigure[][]{
\label{fig:exc-a}
\includegraphics[height=0.8in]{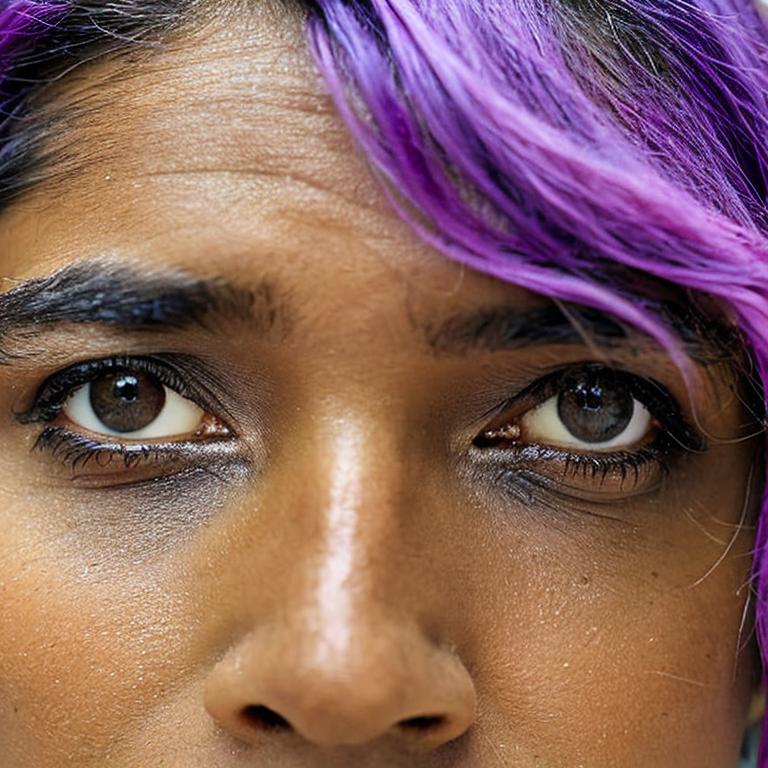}}
\subfigure[][]{
\label{fig:exc-b}
\includegraphics[height=0.8in]{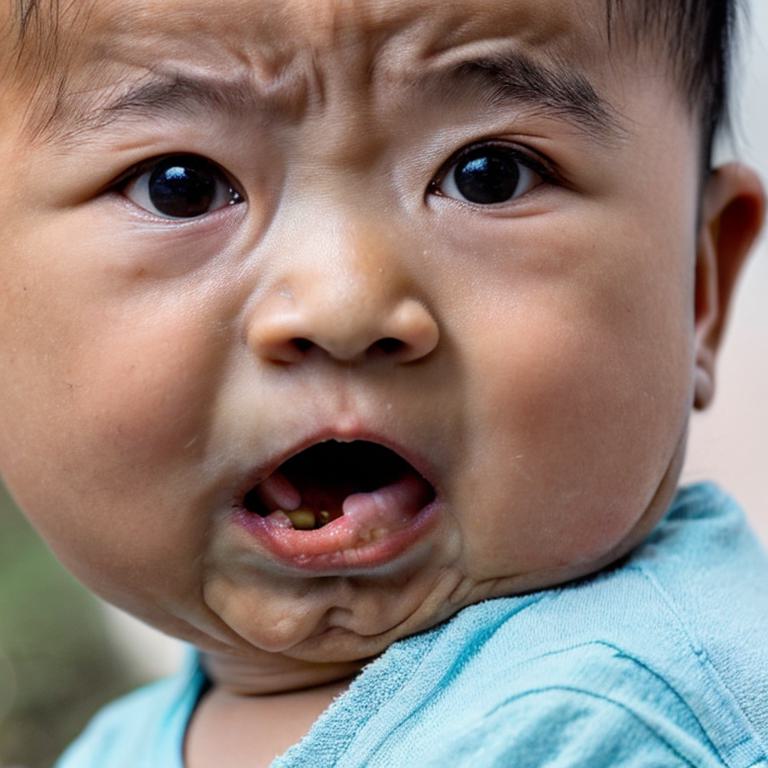}}
\subfigure[][]{
\label{fig:exc-c}
\includegraphics[height=0.8in]{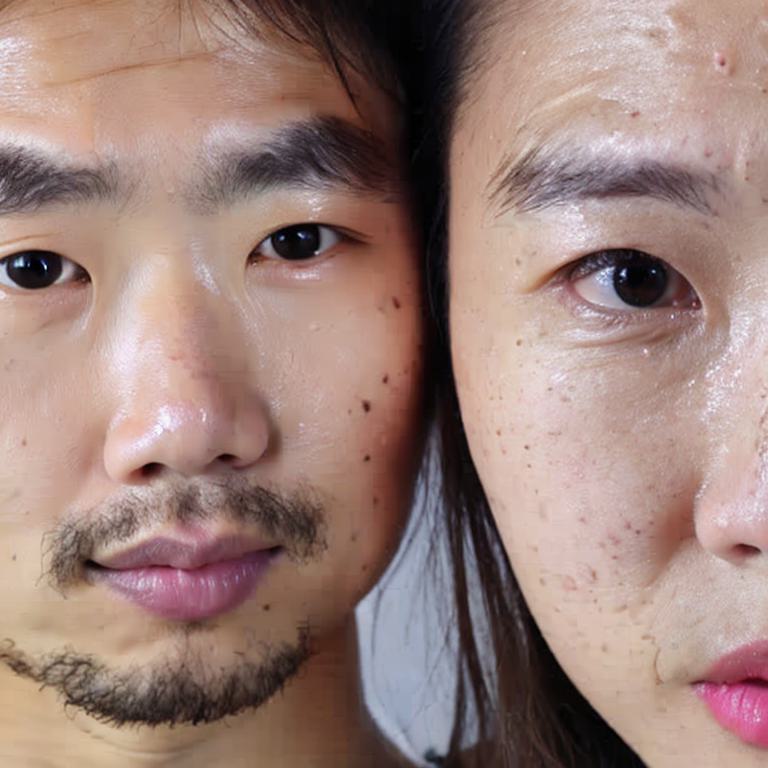}}
\caption[Excluded generated images]{Generated images that were not selected for the final dataset:
\subref{fig:exc-a} the face is very zoomed making it difficult to recognize,  
\subref{fig:exc-b} the mouth of the generated face is badly shaped and does not seem realistic,
\subref{fig:exc-c} the generated image contains two faces instead of one.}
\label{fig:excluded}
\vspace{-4mm}
\end{figure}

The former process of prompting and applying the stable diffusion model must be repeated as many times as necessary to obtain the final face image dataset. Regarding the number of trials with the same prompt, we determined that repeating an experiment more than $20$ times does not increase image quality. 
Naturally, not every image that is generated in each phase meets the predetermined standards. As a result, the images underwent manual inspection and filtering to ensure their high quality. The criteria used for filtering stemmed from the constraints imposed during the prompt formulation process. Figure~\ref{fig:excluded} shows some examples of images that were excluded from the final dataset.

The final version of the \textit{Stable Diffusion Face-image Dataset (SDFD)} consists of $1000$ images accompanied with the relevant information. More specifically, it includes the name of each image, along with the corresponding prompt and the time needed to generate the image. We also keep the labels of all the attributes used in each prompt.
\textit{SDFD} is publicly available for research use  \href{https://drive.google.com/file/d/1TVbnxCNCfzOL7GeG-7gUvbYruxKzlZUE/view?usp=sharing}{online} .

\section{DATASET EVALUATION AND COMPARISON ON ATTRIBUTE CLASSIFICATION}

Since FairFace \cite{karkkainen2021fairface} is a face dataset that also aims to reduce bias, we decided to compare it with our generated face dataset (\textit{SDFD}). We also chose the LFW dataset \cite{huang2008labeled} to compare with \textit{SDFD} since it constitutes a real face image dataset that spans the range of conditions typically encountered in everyday life, while it serves as a widely recognized benchmark for face analysis research. The comparison was performed in terms of both image classification and image distribution in the space, with regard to specific attributes, to conclude if our suggested dataset covers a wider range of face characteristics. 

Concerning the image classification task, we trained a ResNet-18 network using two publicly available face image datasets \textit{CelebA dataset} \cite{liu2015faceattributes} and \textit{UTKface} \cite{zhang2017age}, and then used \textit{SDFD}, FairFace and LFW to infer gender or age as the target attribute. We also conducted experiments using these attributes as well as the race attribute as protected ones, by ensuring that each class of the protected attribute has sufficient representation in each dataset. We have to note that regarding gender classification we excluded from our generated images of \textit{SDFD} the ones with gender label different from male/female. This was done to allow a fair comparison with FairFace and LFW that only include images of these two genders. For the same reason, we labeled \textit{SDFD} images with young/non-young classes.
Furthermore, we evaluated the three datasets in terms of certain metrics: accuracy (ACC), precision (PR), recall (R), F1 score (F1), standard deviation (StD), and skewed error ratio (SkER) computed as the ratio of (max group error)/(min group error). These were computed for each individual class separately. Table~\ref{tab:metrics} shows the corresponding results. In each row, the dataset that appears to have the worst performance, i.e. is the most challenging, is highlighted in bold. As can be seen, \textit{SDFD} fared worse than FairFace and LFW in both gender and age classification in two of the four cases. We can attribute the lower performance of \textit{SDFD} to its higher diversity, or its higher divergence from the datasets used for training, compared with FairFace and LFW. 

\begin{table}
  \centering
  \caption{Metrics evaluation on image classification with target attributes (gender, age) and SDFD, FairFace and LFW datasets.}
  \label{tab:metrics}
  \begin{tabular}{ p{0.9cm}  p{0.9cm}  p{0.7cm}  p{0.4cm} p{0.4cm} p{0.4cm} p{0.4cm} p{0.4cm} p{0.5cm}}
    \hline
    \textbf{Dataset}   & \textbf{Trained on} & \textbf{Target} & \textbf{ACC} & \textbf{PR} & \textbf{R} & \textbf{F1} & \textbf{StD} & \textbf{SkER}  \\
    \hline
\textbf{SDFD}     & CelebA  & Gender & 0.70 & 0.65                        & 0.87 & 0.74 & 0.52 & 3.70  \\
FairFace                        & CelebA  & Gender & 0.75                        & 0.72 & 0.86 & 0.78 & 0.49 & 2.52 \\ 
LFW                        & CelebA  & Gender & 0.90                        & 0.94 & 0.94 & 0.94 & 0.31 & 1.07 \\ \hline
SDFD                            & UTKface & Gender & 0.70                        & 0.84                        & 0.51 & 0.63 & 0.51 & 4.94 \\
\textbf{FairFace} & UTKface & Gender & 0.58                        & 0.65                        & 0.25 & 0.36 & 0.58 & 5.45 \\ 
LFW & UTKface & Gender & 0.73                        & 0.40                        & 0.40 & 0.40 & 0.52 & 1.02 \\\hline
\textbf{SDFD}     & CelebA  & Age    & 0.54                        & 0.39                        & 0.70 & 0.50 & 0.62 & 3.71 \\
FairFace                        & CelebA  & Age    & 0.68                        & 0.82                        & 0.71 & 0.76 & 0.56 & 1.92 \\
LFW                        & CelebA  & Age    & 0.58                        & 0.24                        & 0.79 & 0.37 & 0.54 & 12.16 \\ \hline
SDFD    & UTKface & Age    & 0.55                        & 0.40                        & 0.80 & 0.54 & 0.59 & 5.93 \\
FairFace                        & UTKface & Age    & 0.60                        & 0.74                        & 0.69 & 0.71 & 0.63 & 1.25 \\
\textbf{LFW}                        & UTKface & Age    & 0.43                        & 0.36                        & 0.86 & 0.51 & 0.59 & 10.86 \\ \hline
  \end{tabular}
\end{table}


According to Table~\ref{tab:metrics}, it appears that using the UTKface for training results in lower performance than CelebA, in both gender and age classification. 

Table~\ref{tab:protected} presents the accuracy of classifying images from \textit{SDFD}, FairFace and LFW using either gender as target attribute and age as the protected one, or vice versa. In each row, the dataset with the poorest performance is highlighted in bold. In this case, the findings demonstrate that \textit{SDFD} and LFW perform worse in age classification but when the target attribute is gender FairFace performs worse.


\begin{table}
  \centering
  \caption{Accuracy evaluation on image classification with both target and protected attributes (gender, age) and SDFD, FairFace and LFW datasets.}
  \label{tab:protected}
  \begin{tabular}{ p{1.1cm}  p{1.4cm}  p{1.1cm}  p{2.1cm} p{1cm} }
    \hline
    \textbf{Dataset} & \textbf{Trained on} & \textbf{Target} & \textbf{Protected} & \textbf{ACC} \\
    \hline
SDFD     & CelebA  & Gender & Age (young)     & 0.77 \\
\textbf{FairFace} & CelebA  & Gender & Age (young)     & 0.70 \\ 
LFW     & CelebA  & Gender & Age (young)     & 0.89 \\ \hline
SDFD     & UTKface & Gender & Age (young)     & 0.77 \\
\textbf{FairFace} & UTKface & Gender & Age (young)     & 0.54 \\ 
LFW      & UTKface & Gender & Age (young)     & 0.59 \\ \hline
\textbf{SDFD}     & CelebA  & Age    & Gender (female) & 0.46 \\
FairFace & CelebA  & Age    & Gender (female) & 0.70 \\ 
LFW & CelebA  & Age    & Gender (female) & 0.77 \\ \hline
\textbf{SDFD}     & UTKface & Age    & Gender (female) & 0.38 \\
FairFace & UTKface & Age    & Gender (female) & 0.58 \\ 
LFW & UTKface & Age    & Gender (female) & 0.50 \\ \hline
SDFD     & CelebA  & Age    & Gender (male)   & 0.60 \\ 
FairFace & CelebA  & Age    & Gender (male)   & 0.66 \\ 
\textbf{LFW} & CelebA  & Age    & Gender (male)   & 0.58 \\\hline
SDFD     & UTKface & Age    & Gender (male)   & 0.67 \\
FairFace & UTKface & Age    & Gender (male)   & 0.55 \\
\textbf{LFW} & UTKface  & Age    & Gender (male)   & 0.33 \\ \hline
  \end{tabular}
  \vspace{-5mm}
\end{table}

Table~\ref{tab:protected} shows that UTKface continues to lead to lower performance compared with CelebA in terms of gender classification. Regarding the two genders examined, \textit{SDFD} has lower accuracy in female prediction, while FairFace and LFW have the opposite effect. Again, CelebA dataset leads to better accuracy compared with UTKface when using age as the target attribute.

Table~\ref{tab:raceProt} presents the accuracy of classifying images from \textit{SDFD}, FairFace and LFW using either gender or age as target attribute and race as the protected one. The Pacific Islander is a race not included in the FairFace dataset, thus there are dashes in the corresponding table cells. The same happens for the LFW for races other than Black, Indian and White. In each row, the dataset that appears to have the worst performance is highlighted in bold. In this case, the findings demonstrate that \textit{SDFD} performs the worst in most cases.


\begin{table*}
\small
  \centering
  \caption{Accuracy on image classification with both target (gender, age) and protected (race) attributes on SDFD, FairFace and LFW datasets (PI: Pacific Islander, ME: Middle Eastern, SEA: South-east Asian, EA: East Asian).}
  \label{tab:raceProt}
  \small
  \resizebox{\textwidth}{!}{\begin{tabular}{ p{1.1cm}  p{1.5cm}  p{1cm}  p{1.4cm} p{0.8cm} p{0.8cm} p{0.8cm} p{0.8cm} p{0.8cm} p{0.8cm} p{0.8cm} p{0.8cm}}
    \hline
    \textbf{Dataset}  & \textbf{Trained on} & \textbf{Target} & \textbf{Protected} & \multicolumn{8}{c}{\textbf{Accuracy}}  \\
     &
   &
   &
   &
  Black &
  Indian &
  PI & 
  ME & 
  White &
  Latino &
  SEA & 
  EA \\ 
    \hline
\textbf{SDFD}     & CelebA  & Gender & Race & 0.65 & 0.70  & 0.70  & 0.77 & 0.66 & 0.57 & 0.74 & 0.78 \\
 
FairFace & CelebA  & Gender & Race & 0.67 & 0.72 & -    & 0.81 & 0.77 & 0.78 & 0.77 & 0.74 \\
LFW & CelebA  & Gender & Race & 0.84 & 0.92 & -    & - & 0.91 & - & - & - \\ \hline

SDFD     & UTKface & Gender & Race & 0.65 & 0.70  & 0.59 & 0.63 & 0.72 & 0.70  & 0.70  & 0.69 \\
 
\textbf{FairFace} & UTKface & Gender & Race & 0.52 & 0.57 & -    & 0.69 & 0.59 & 0.51 & 0.58 & 0.55 \\

LFW & UTKface  & Gender & Race & 0.72 & 0.71 & -    & - & 0.73 & - & - & - \\ \hline
 
\textbf{SDFD}     & CelebA  & Age    & Race & 0.57 & 0.50  & 0.55 & 0.67 & 0.52 & 0.62 & 0.53 & 0.52 \\

FairFace & CelebA  & Age    & Race & 0.62 & 0.66 & -    & 0.66 & 0.69 & 0.68 & 0.69 & 0.74 \\
LFW & CelebA  & Age & Race & 0.72 & 0.71 & -    & - & 0.73 & - & - & - \\ \hline

\textbf{SDFD}     & UTKface & Age    & Race & 0.64 & 0.54 & 0.45 & 0.50  & 0.70  & 0.54 & 0.37 & 0.62 \\
 
FairFace & UTKface & Age    & Race & 0.55 & 0.57 & -    & 0.57 & 0.58 & 0.59 & 0.63 & 0.67 \\
LFW & UTKface  & Age & Race & 0.72 & 0.71 & -    & - & 0.73 & - & - & - \\ \hline
  \end{tabular}}
\end{table*}

Table~\ref{tab:raceProt} shows that in gender classification using race as a protected attribute, UTKface performs worse than CelebA. \textit{SDFD} performs poorly in the \textit{Latino} race with CelebA training and the \textit{Pacific Islander} race with UTKface. FairFace underperforms in the \textit{Black} race, regardless of the dataset used for training. In terms of age classification using race as a protected attribute, no training dataset outperforms the other. \textit{SDFD} shows low performance in the \textit{Indian} race with CelebA training and the \textit{Southeast Asian} race with UTKface. FairFace performs poorly in the \textit{Black} race again, independent of the dataset used during training. LFW has the lowest performance in the \textit{Indian} race.

A tentative observation based on the previous results suggests that CelebA and UTKface may not be sufficient for training a model that generalizes well. This could explain the low accuracy values observed during cross-dataset validation. It is worth noting that these datasets contain images with limited diversity, which might contribute to these outcomes. Furthermore, a general observation on the experimental findings demonstrates that \textit{SDFD} proved to be equally challenging for image classification compared with FairFace and LFW, but it is much smaller in size, making it easier and less costly to use. 

\begin{figure*}
\centering
\subfigure[][SDFD - FairFace]{
\label{fig:tsne1}
\includegraphics[height=1.3in]{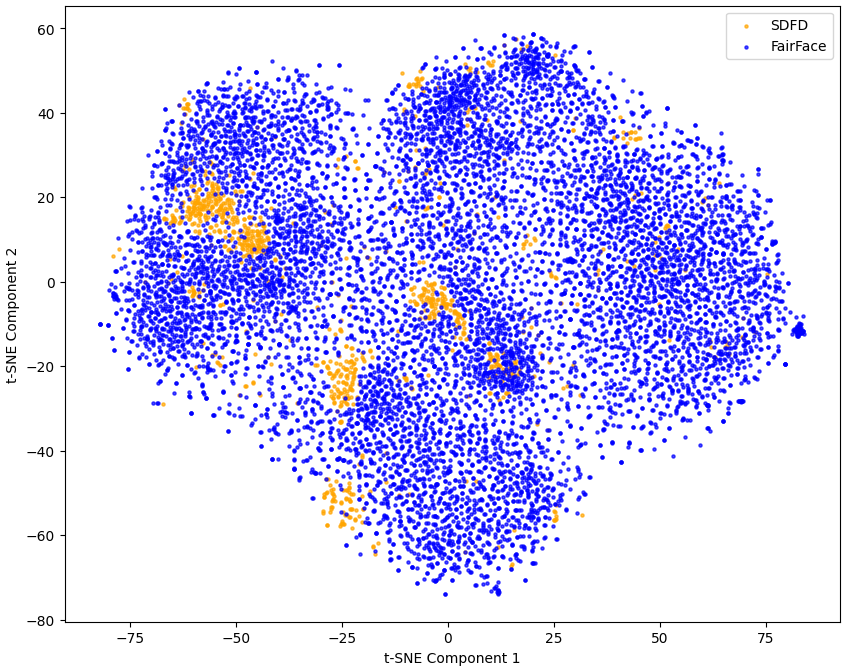}}
\subfigure[][SDFD - LFW]{
\label{fig:tsne2}
\includegraphics[height=1.3in]{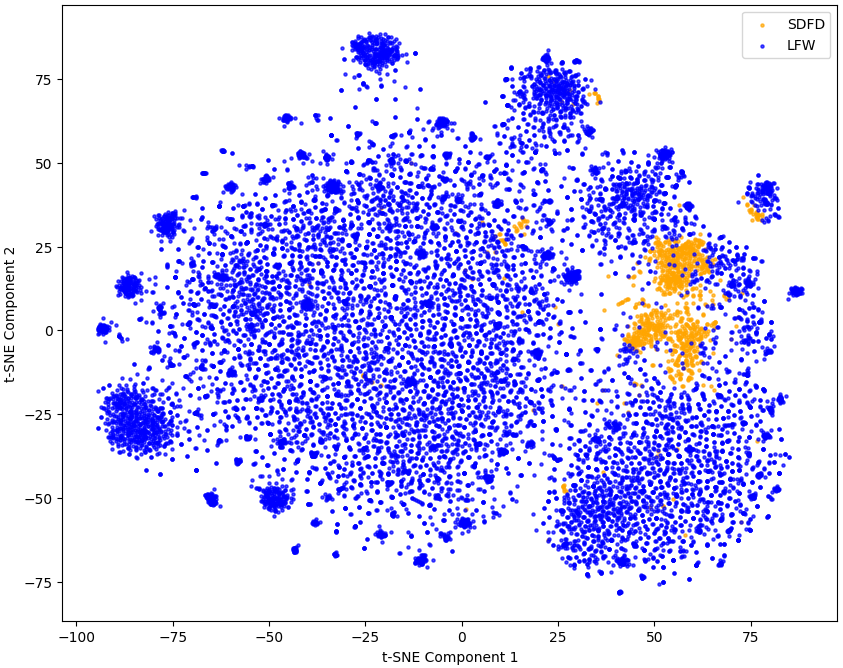}}
\subfigure[][SDFD - FairFace (Facer)]{
\label{fig:tsne3}
\includegraphics[height=1.3in]{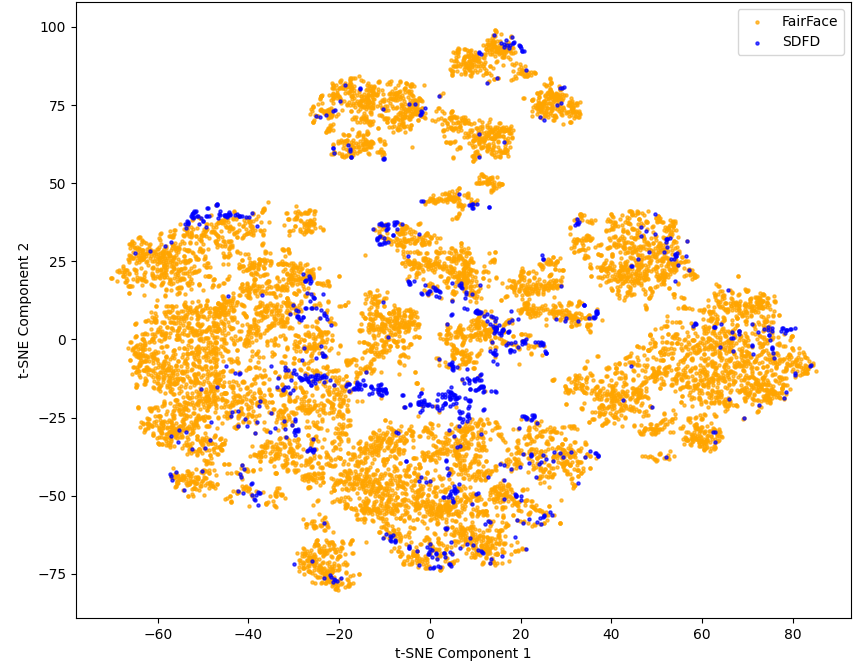}}
\subfigure[][SDFD - LFW (Facer)]{
\label{fig:tsne4}
\includegraphics[height=1.3in]{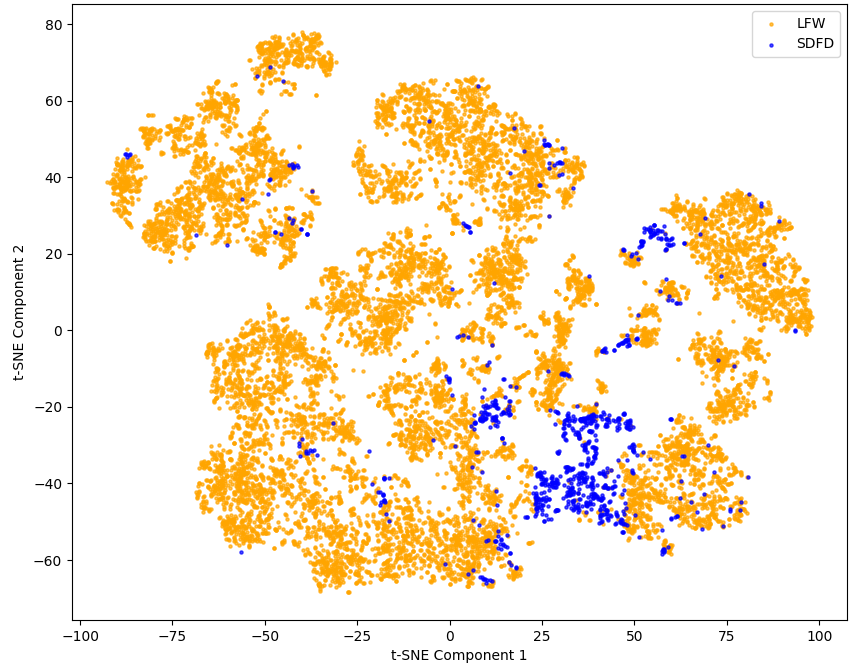}}
\caption[t-SNE visualization]{t-SNE visualization of embeddings from visual features (\subref{fig:tsne1}, \subref{fig:tsne2}) and Facer attributes (\subref{fig:tsne3}, \subref{fig:tsne4}) of SDFD versus FairFace  and LFW respectively.
}
\label{fig:tsne}
\vspace{-6mm}
\end{figure*}

Aside from comparing the three datasets using image classification, we also visualize their images in two dimensions to analyze their spatial distribution. 
Figure~\ref{fig:tsne} illustrates the t-SNE visualization \cite{ljpvd2008visualizing} of \textit{SDFD}, FairFace and LFW datasets. More precisely, in Figure~\ref{fig:tsne1} it appears that our suggested dataset, \textit{SDFD}, fills in some of the FairFace's gaps. Considering the sizes of the two datasets—\textit{SDFD} has $1000$ images, while FairFace has $10,953$—the former observation appears to be noteworthy. For efficiency considerations, we decided to compare only the FairFace dataset's validation set consisting of $10,953$ images rather than the entire collection ($108,501$ images).

Figure~\ref{fig:tsne2} displays the t-SNE visualization \cite{ljpvd2008visualizing} of \textit{SDFD} and LFW dataset. It appears also here that our suggested dataset, \textit{SDFD}, fills in some of LFW's gaps. Considering the sizes of the two datasets—\textit{SDFD} has $1000$ images, while LFW has $14,089$—the former appears to be noteworthy.
Additionally, we extracted a set of 40 facial attributes from the two datasets using the \textit{Facer framework} \cite{zheng2022farl}. 
These attributes are listed in Table~\ref{tab:attrs}.
We next utilized t-SNE visualization to examine the dispersion of \textit{Facer} characteristics across the datasets. Figure~\ref{fig:tsne3} illustrates the t-SNE visualization of face attributes of FairFace and \textit{SDFD}, whereas Figure~\ref{fig:tsne4} illustrates LFW and \textit{SDFD}. In both plots, we can see that, despite its smaller size compared to the FairFace and LFW datasets, \textit{SDFD} covers a large attribute area. The former is more obvious in the plot of FairFace.


\begin{table}
\small
\centering
\caption{Face attributes extracted using \textit{Facer} framework.}
  \label{tab:attrs}
    \begin{tabular}{rrrrrr}
      \hline
      \bf \# & \bf  Attribute Name & \bf \# & \bf  Attribute Name\\
      \hline
      1      & 5 o' Clock Shadow   &  21      & Male    \\
      2      & Arched Eyebrows     &  22      & Mouth Slightly Open   \\
      3      & Attractive          &  23      & Mustache   \\
      4      & Bags Under Eyes     &  24      & Narrow Eyes   \\
      5      & Bald                &  25      & No Beard    \\
      6      & Bangs               &  26      & Oval Face    \\
      7      & Big Lips            &  27      & Pale Skin \\
      8      & Big Nose            &  28      & Pointy Nose \\
      9      & Black Hair          &  29      & Receding Hairline \\
      10     & Blond Hair          &  30      & Rosy Cheeks \\
      11     & Blurry              &  31      & Sideburns \\
      12     & Brown Hair          &  32      & Smiling \\
      13     & Bushy Eyebrows      &  33      & Straight Hair \\
      14     & Chubby              &  34      & Wavy Hair \\
      15     & Double Chin         &  35      & Wearing Earrings \\
      16     & Eyeglasses          &  36      & Wearing Hat \\
      17     & Goatee              &  37      & Wearing Lipstick \\
      18     & Gray Hair           &  38      & Wearing Necklace \\
      19     & Heavy Makeup        &  39      & Wearing Necktie \\
      20     & High Cheekbones     &  40      & Young \\
      \hline
    \end{tabular}
    \vspace{-6mm}
\end{table}


Diversity and inclusion are crucial features of the proposed dataset that may not be clearly valued but constitute important characteristics of \textit{SDFD}.


\section{DISCUSSION}

The presented face image generation approach produced a realistic and high-quality dataset. However, there were several complications encountered during this process. More specifically, there were some terms describing attributes that were not possible or hard to be applied in the final images. Table~\ref{tab:limitations} summarizes these problems along with their corresponding terms. The training datasets of the stable diffusion model could likely be the source of this issue. The former demonstrates the lack of attribute variety in available face image datasets. Although some of the words specified in Table~\ref{tab:limitations} may be too particular to be shown in a generated image, e.g. the religious items, the majority of them should be applicable. Figure~\ref{fig:excluded} shows some examples of the above limitations.


\begin{table}
\centering
  \caption{Attribute problematic cases and their corresponding terms.}
  \label{tab:limitations}
\small
    \begin{tabular}{p{4cm} p{4cm}}
      \hline
      \bf  Attribute Problem & \bf  Attribute Terms\\
      \hline
      Terms unable to be applied on result image           & angry, eyepatch, scar, pimples, freckles, teeth braces, teeth brackets, lips-botox, face lifting, eyeshadow, shayla, shpitzel, sheitel, zuchetto, mitre, chin in hand pose, hands up pose, hands-on-waist pose    \\ \hline
      Terms needing many iterations to be applied   &   tattoo\\ \hline
      Terms that are not properly applied &   excited, sunglasses, non-binary terms (depict two people), color mismatching, eyes' color/wrinkles result in zoomed face, long prompts 
      \\ \hline
      Combinations of terms suffering from above issues 
      &  heavy makeup+baby, beard+baby, Indian+helmet, bald+helmet \\ 
       
      \hline
    \end{tabular}
    \vspace{-6mm}
\end{table}

\begin{figure}
\centering
\subfigure[][]{
\label{fig:str-a}
\includegraphics[height=1in]{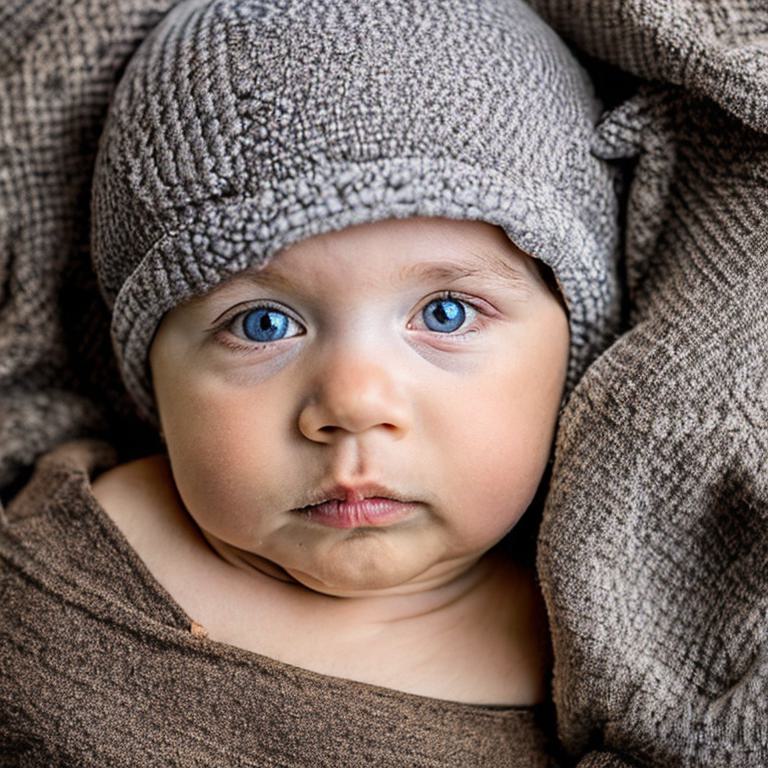}}
\subfigure[][]{
\label{fig:str-b}
\includegraphics[height=1in]{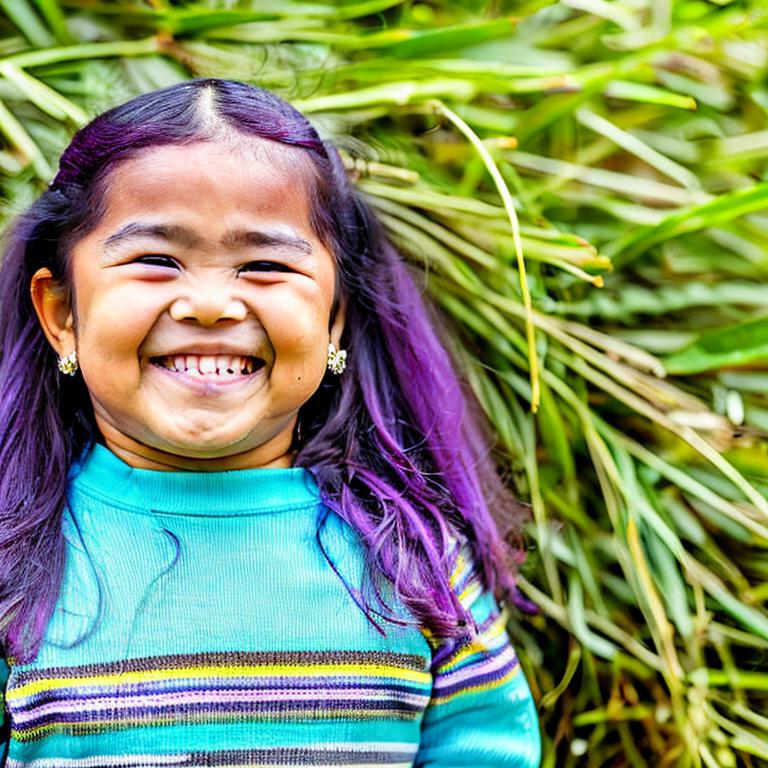}}
\subfigure[][]{
\label{fig:str-c}
\includegraphics[height=1in]{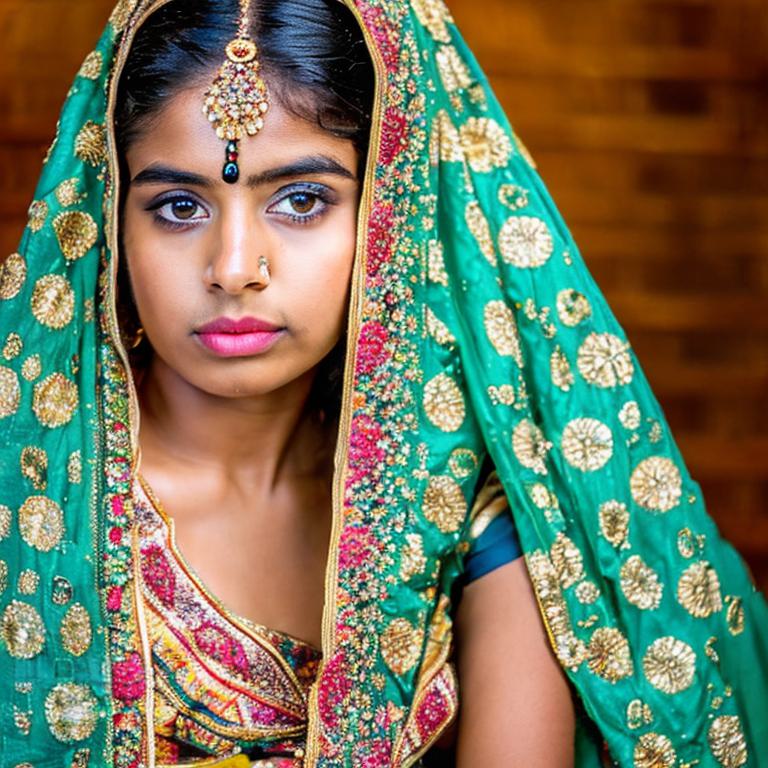}}
\caption[Stereotypical generated images]{Generated images that might reinforce race stereotypes:
\subref{fig:str-a} the prompt given was a Black baby boy with blue eyes. However a White baby was generated,  
\subref{fig:str-b} a Pacific Islander girl with leafy background, 
\subref{fig:str-c} an Indian girl wearing traditional clothes.}
\label{fig:stereotypicals}
\vspace{-5mm}
\end{figure}

Bias in the generated images was another problem that surfaced. Particularly when specific sentences were used in prompting, stereotypical patterns appeared in the resulting images. For instance, when the race of Pacific Islander was used, most of the times the background of the resulting image was one of foliage. Besides, concerning the Indian race, the majority of generated images depicted people with colorful, traditional attire and jewellery. In other cases, the combination of Black race and colorful eyes mostly resulted in White people with colorful eyes. Figure~\ref{fig:stereotypicals} shows some examples of images that were not selected for the final dataset due to perpetuating stereotypes. Despite our efforts in the opposite direction, the introduced dataset may contain various recognized biases or societal prejudices.

The dataset evaluation and comparison for attribute classification revealed that \textit{SDFD} gives comparable results to the FairFace and LFW datasets for the metrics under consideration. In particular, none of the three datasets performed consistently well. However, in most cases, \textit{SDFD} proved to be equally or even more challenging in image classification, with the obvious benefit of its substantially smaller size. The former makes \textit{SDFD} easier to use as an evaluation dataset.

\section{CONCLUSIONS AND FUTURE WORK}

In this work, we present a general methodology to generate synthetic face image datasets, which can be adapted according to the context of use. Following the steps and the specifications outlined, realistic and diversified face image datasets may be created. Furthermore, we introduce a novel face image dataset (\textit{SDFD}), characterized by diversity and inclusiveness. SDFD may be utilized as an evaluation set for models that perform demographic attribute prediction. Including an evaluation set in such tasks facilitates the generalization of models to diverse populations, enhancing their applicability in real-world contexts. When we compared our suggested produced dataset to existing ones, we noticed that our approach made it equally or even more challenging to classify images based on gender and/or age, while being significantly smaller in size, and therefore easier and less costly to use. Although the size of \textit{SDFD} is relatively small, it sufficiently covers the diversity implied by the utilized attributes. Besides, a two-dimensional visualization of the datasets revealed that \textit{SDFD} has a very good spatial dispersion. The former might imply that \textit{SDFD} is a more inclusive dataset with higher face attribute variety. In the future, we are intend to experiment with more generative models and expand the scale of our dataset, while also including even more facial attributes such as scars, braces, and other features that we were unable to integrate in the current dataset version. Finally, we aim to experiment with different traits like disabilities and disfigurements in order to include more underrepresented groups.

\section{ACKNOWLEDGMENTS}

This research work was funded by the European Union under the Horizon Europe MAMMOth project, Grant Agreement ID: 101070285.


{\small
\balance
\renewcommand{\citepunct}{,\penalty\citepunctspace}
\bibliographystyle{ieee}
\bibliography{SDFD}
}

\end{document}